\pdfoutput=1

\documentclass[11pt]{article}

\usepackage{ACL2023}

\usepackage{times}
\usepackage{latexsym}
\usepackage{amsmath}
\usepackage{graphicx}
\usepackage{makecell}
\usepackage{booktabs}
\usepackage{enumitem}
\usepackage{booktabs}
\usepackage{multirow}
\usepackage{stfloats}
\usepackage{cleveref}
\usepackage{bbding} 

\usepackage[T1]{fontenc}

\usepackage[utf8]{inputenc}

\usepackage{microtype}

\usepackage{inconsolata}
\usepackage{fontawesome}

\newcommand{\sist}{\textsuperscript{1}}
\newcommand{\vdi}{\textsuperscript{2}}
\newcommand{\damo}{\textsuperscript{3}}

\title{Do PLMs Know and Understand Ontological Knowledge?}

\author{   
    \textbf{Weiqi Wu}$^{1}$,
    \textbf{Chengyue Jiang}$^{1,2}$,
    \textbf{Yong Jiang}$^{3\ast}$,
    \textbf{Pengjun Xie}$^{3}$,
    \textbf{Kewei Tu}$^{1,2}$\thanks{$~~$Yong Jiang and Kewei Tu are corresponding authors.} \\
    \sist School of Information Science and Technology, ShanghaiTech University \\
    \vdi Shanghai Engineering Research Center of Intelligent Vision and Imaging \\
    \damo DAMO Academy, Alibaba Group, China \\
    \texttt{\{wuwq,jiangchy,tukw\}@shanghaitech.edu.cn}\\
    \texttt{\{yongjiang.jy,chengchen.xpj\}@alibaba-inc.com }
}

\begin{document}
\maketitle
\begin{abstract}
    Ontological knowledge, which comprises classes and properties and their relationships, is integral to world knowledge. It is significant to explore whether Pretrained Language Models (PLMs) know and understand such knowledge. However, existing PLM-probing studies focus mainly on factual knowledge, lacking a systematic probing of ontological knowledge. In this paper, we focus on probing whether PLMs store ontological knowledge and have a semantic understanding of the knowledge rather than rote memorization of the surface form.
    To probe whether PLMs know ontological knowledge, we investigate how well PLMs memorize: (1) types of entities; (2) hierarchical relationships among classes and properties, e.g., \textit{Person} is a subclass of \textit{Animal} and \textit{Member of Sports Team} is a subproperty of \textit{Member of}; (3) domain and range constraints of properties, e.g., the subject of \textit{Member of Sports Team} should be a \textit{Person} and the object should be a \textit{Sports Team}. To further probe whether PLMs truly understand ontological knowledge beyond memorization, we comprehensively study whether they can reliably perform logical reasoning with given knowledge according to ontological entailment rules. Our probing results show that PLMs can memorize certain ontological knowledge and utilize implicit knowledge in reasoning. However, both the memorizing and reasoning performances are less than perfect, indicating incomplete knowledge and understanding.
\end{abstract}

\section{Introduction}
Pretrained Language Models (PLMs) have orchestrated impressive progress in NLP across a wide variety of downstream tasks, including knowledge-intensive tasks. Previous works propose that PLMs are capable of encoding a significant amount of knowledge from the pretraining corpora~\cite{AlKhamissi2022ARO}, and determine to explore the kinds of knowledge within PLMs. Existing probing works mainly focus on factual knowledge associated with instances~\cite{petroni-etal-2019-language, jiang-etal-2020-know, safavi-koutra-2021-relational}. Meanwhile, although classes (concepts) have raised some research interest~\cite{Bhatia2020TransformerNO, peng2022copen, lin-ng-2022-bert}, there is no systematic study of ontological knowledge.

\begin{figure}[t]
    \centering
    \includegraphics[width=0.47\textwidth]{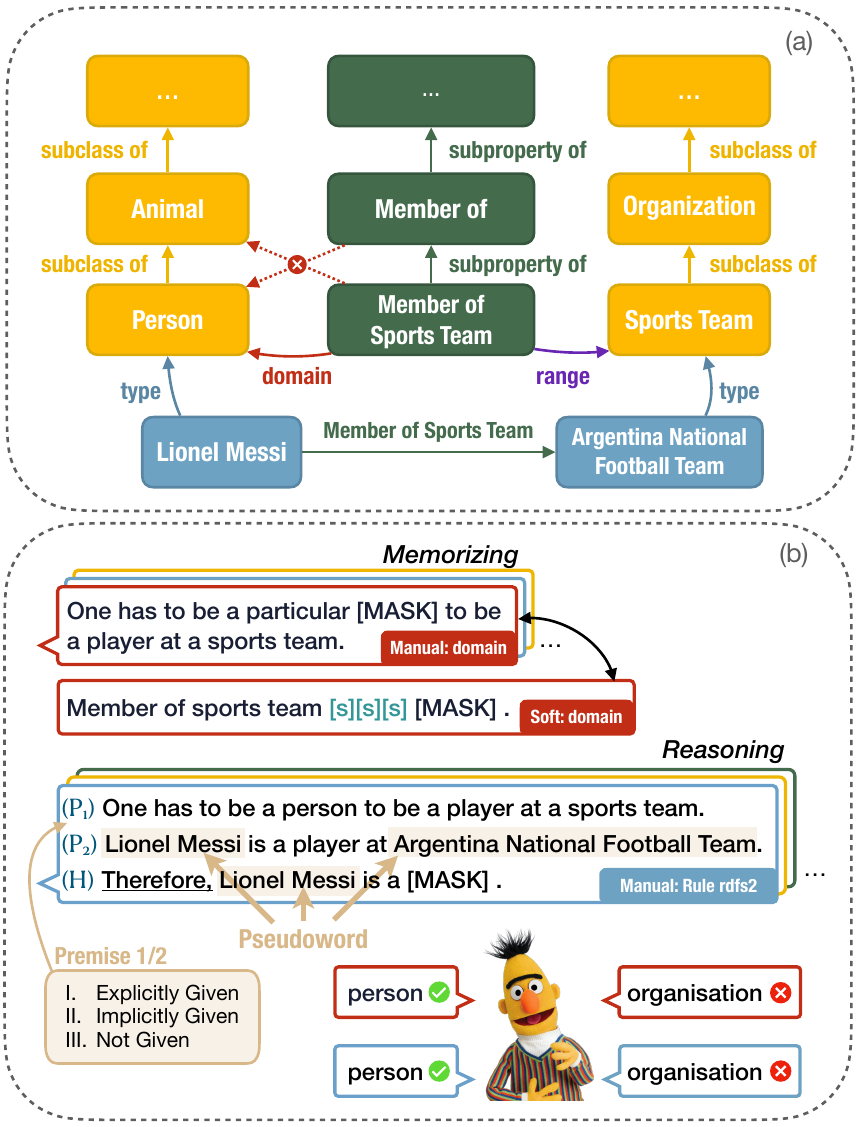}
    \caption{(a) An example of an ontological knowledge graph. (b) Potential manual and soft prompts to probe the knowledge and corresponding semantics. Instances are replaced by pseudowords in reasoning experiments to mitigate potential interference from model memory.}
    \label{f1}
\end{figure}

Ontological knowledge models the world with a set of classes and properties and the relationships that hold between them~\cite{Nilsson2006OntologicalCF, Kumar2019AnOD}. It plays a vital role in many NLP tasks such as question answering by being injected into~\cite{goodwin-demner-fushman-2020-enhancing} or embedded outside deep neural networks~\cite{8047276}. Therefore, it is essential to explore whether PLMs can encode ontological knowledge and have a semantic understanding of the knowledge rather than rote memorizing its surface form. 

In this paper, we first probe PLM's memorization of ontological knowledge. Specifically, as shown in Figure~\ref{f1}(a), we construct memorization tests about (1) Types of entities. Entities can be categorized into classes, as Lionel Messi is a \textit{Person} and Argentina National Football Team is a \textit{Sports Team}. (2) Hierarchical relationships between classes, e.g., \textit{Person} is a subclass of \textit{Animal}. (3) Hierarchical relationships between properties, e.g., \textit{Member of Sports Team} is a subproperty of \textit{Member of}. (4) Domain constraints of properties. It specifies information about the subjects to which a property applies. For example, the subject of \textit{Member of Sports Team} should be an instance of \textit{Person}. (5) Range constraints of properties. Similar to domain, range specifies information about the object of a property, such as the object of \textit{Member of Sports Team} should be an instance of \textit{Sports Team}. Experiments prove that PLMs store a certain amount of ontological knowledge.

To further examine whether PLMs understand ontological knowledge, we investigate if PLMs can correctly perform logical reasoning that requires ontological knowledge. Illustrated in Figure~\ref{f1}(b), given the fact triple (Lionel Messi, \textit{Member of Sports Team}, Argentina National Football Team) along with property constraints, we can perform type inferences to conclude that Lionel Messi is a \textit{Person}, and Argentina National Football Team is a \textit{Sports Team}. We comprehensively investigate the reasoning capability of PLMs over ontological knowledge following six entailment rules. Experiments show that PLMs can apply implicit ontological knowledge to draw conclusions through reasoning, but the accuracy of their reasoning falls short of perfection. This observation suggests that PLMs possess a limited understanding of ontological knowledge.

In summary, we systematically probe whether PLMs know and understand ontological knowledge. Our main contributions can be summarized as follows: (1) We construct a dataset that evaluates the ability of PLMs to memorize ontological knowledge and their capacity to draw inferences based on ontological entailment rules. (2) We comprehensively probe the reasoning ability of PLMs by carefully classifying how ontological knowledge is given as a premise. (3) We find that PLMs can memorize certain ontological knowledge but have a limited understanding. We anticipate that our work will facilitate more in-depth research on ontological knowledge probing with PLMs. The code and dataset are released at \url{https://github.com/vickywu1022/OntoProbe-PLMs}.
\section{Benchmark Construction}

In this section, we present our methodology for ontology construction and the process of generating memorizing and reasoning tasks based on the ontology for our probing analysis.

\subsection{Ontology Building}
\label{sec:ontology build}
\paragraph{Class} We use DBpedia~\cite{Auer2007DBpediaAN} to obtain classes and their instances. Specifically, we first retrieve all 783 classes in DBpedia, then use SPARQL~\cite{hommeaux2011SPARQLQL} to query their instances using the \texttt{type} relation and superclasses using the \texttt{subclass-of} relation. We sample 20 instances for each class.

\paragraph{Property} Properties are collected based on DBpedia and Wikidata~\cite{Vrandei2014WikidataAF} using the following pipeline: (1) Obtain properties from Wikidata and use \textit{subproperty of (P1647)} in Wikidata to find their superproperties. (2) Query the domain and range constraints of the properties using \textit{property constraint (P2302)} in Wikidata. (3) Align the Wikidata properties with DBpedia properties by \textit{equivalent property (P1628)}. (4) Query the domain and range constraints of the properties in DBpedia. (5) Cleanse the collected constraints using the above-collected class set as vocabulary. We choose 50 properties with sensible domain, range and superproperties.

\subsection{Construction of Memorizing Task}
The memorizing task consists of five subtasks, each probing the memorization of an ontological relationship: (1) \textbf{TP}: types of a given instance, (2) \textbf{SCO}: superclasses of a given class, (3) \textbf{SPO}: superproperties of a given property, (4) \textbf{DM}: domain constraint on a given property, and (5) \textbf{RG}: range constraint on a given property. Every subtask is formulated as a cloze-completion problem, as shown in Figure~\ref{f1}(b). Multiple correct answers exist for TP, SCO, and SPO, which form a chain of classes or properties. There is only one correct answer for DM and RG, as it is not sound to declare an expanded restriction on a property. For instance, \textit{Animal} is too broad as the domain constraint of the property \textit{Member of Sports Team (P54)}, hence applying \textit{Person} as the domain. 

We construct the dataset for each subtask using the ontology built in Sec.~\ref{sec:ontology build} and reserve 10 samples for training and 10 for validation to facilitate few-shot knowledge probing. The statistics of the dataset for each subtask are shown in Table~\ref{mem}.

\begin{table}[tb]
\centering
\tabcolsep=0.15cm
\scalebox{0.8}{
\begin{tabular}{cccccc}
\toprule[1.5pt]
 \textbf{Task}  & \textbf{Ontological Rel.} & \textbf{Candidate} & \textbf{Train} & \textbf{Dev} & \textbf{Test}\\
\midrule
\textbf{TP} & type &  class & 10 & 10 & 8789 \\
\textbf{SCO} & subclass of &  class & 10 & 10 & 701\\
\textbf{SPO} & subproperty of &  property & 10 & 10 & 39\\
\textbf{DM} & domain & class & 10 & 10 & 30\\
\textbf{RG} & range & class & 10 & 10 & 28\\
\bottomrule[1.5pt]
\end{tabular}}
\caption{Ontological relationship, type of candidate, and dataset size for each memorizing subtask.}
\label{mem}
\end{table}

\subsection{Construction of Reasoning Task}

\begin{table*}[hb]
\centering
\scalebox{0.8}{
\begin{tabular}{cllll}
\toprule[1.5pt]
\textbf{Rule} & \textbf{Premises} & \textbf{Conclusion} & \textbf{Candidate} & \textbf{Remark}\\
\midrule
rdfs2 & \makecell[l]{$[\mathcal{P}_1]$ aaa \texttt{domain} \textbf{\textcolor{orange}{xxx}}.  \\ $[\mathcal{P}_2]$ uuu aaa vvv. } & uuu \texttt{type} \textbf{\textcolor{orange}{xxx}}. & class & \makecell[l]{Type inference through\\domain constraint.} \\
\midrule
rdfs3 & \makecell[l]{$[\mathcal{P}_1]$ aaa \texttt{range} \textbf{\textcolor{orange}{xxx}}. \\ $[\mathcal{P}_2]$ uuu aaa vvv.} & vvv \texttt{type} \textbf{\textcolor{orange}{xxx}}. & class & \makecell[l]{Type inference through\\range constraint.}\\
\midrule
rdfs5 & \makecell[l]{$[\mathcal{P}_1]$ bbb \texttt{subproperty of} \textbf{\textcolor{orange}{ccc}}.  \\ 
$[\mathcal{P}_2]$ aaa \texttt{subproperty of} bbb.} & aaa \texttt{subproperty of} \textbf{\textcolor{orange}{ccc}}. & property & \makecell[l]{Transitivity of \\subproperty.} \\
\midrule
rdfs7 & \makecell[l]{$[\mathcal{P}_1]$ aaa \texttt{subproperty of} \textbf{\textcolor{orange}{bbb}}. \\ $[\mathcal{P}_2]$ uuu aaa vvv.} & uuu \textbf{\textcolor{orange}{bbb}} vvv. & property pattern & \makecell[l]{Property inheritance \\through subproperty.}\\
\midrule
rdfs9 & \makecell[l]{$[\mathcal{P}_1]$ xxx \texttt{subclass of} \textbf{\textcolor{orange}{yyy}}. \\ $[\mathcal{P}_2]$ uuu \texttt{type} xxx.} & uuu \texttt{type} \textbf{\textcolor{orange}{yyy}}. & class & \makecell[l]{Type inheritance \\through subclass.}\\
\midrule
rdfs11 & \makecell[l]{$[\mathcal{P}_1]$ yyy \texttt{subclass of} \textbf{\textcolor{orange}{zzz}}. \\ $[\mathcal{P}_2]$ xxx \texttt{subclass of} yyy.} &  xxx \texttt{subclass of} \textbf{\textcolor{orange}{zzz}}. & class & \makecell[l]{Transitivity of \\subclass.} \\
\bottomrule[1.5pt]
\end{tabular}}
\caption{Entailment rules for the reasoning task. Symbol aaa and bbb represent any random property. Symbols xxx, yyy and zzz represent some classes, and uuu and vvv represent some instances. Constituents of the conclusion highlighted in \textcolor{orange}{orange} are to be masked in the input, and $\mathcal{P}_1$ is the premise that contains the same constituents.}
\label{rules}
\end{table*}

We construct the reasoning task based on the entailment rules specified in the Resource Description Framework Schema (RDFS)\footnote{RDFS is an extension of RDF~\cite{Brickley2002ResourceDF, Gibbins2009ResourceDF}, a widely used and recognized data model. See \url{https://www.w3.org/TR/rdf11-mt/\#rdfs-entailment} for all the entailment rules.}. We propose six subtasks, each probing the reasoning ability following a rule listed in Table~\ref{rules}. For rule rdfs2/3/7, we design a pattern for each property to be used between a pair of instances, e.g., "[X] is a player at [Y] ." for \textit{Member of Sports Team}, where [X] and [Y] are the subject and object, respectively.

Each entailment rule describes a reasoning process: $\mathcal{P}_1 \wedge \mathcal{P}_2 \models \mathcal{H}$, where $\mathcal{P}_1, \mathcal{P}_2$ are the premises and $\mathcal{H}$ is the hypothesis. Similar to the memorizing task, we formulate the reasoning task as cloze-completion by masking the hypothesis (see Figure~\ref{f1}(b)). Premises are also essential to the reasoning process and can be:

\begin{itemize}[leftmargin=*,topsep=5pt]
    \item \textit{Explicitly Given}: The premise is explicitly included in the input of the model, and inferences are made with natural language statements.
    \item \textit{Implicitly Given}: The premise is not explicitly given but memorized by the model as implicit knowledge. The model needs to utilize implicit knowledge to perform inferences, which relieves the effect of context and requires understanding the knowledge. 
    \item \textit{Not Given}: The premise is neither explicitly given nor memorized by the model. It serves as a baseline where the model makes no inference.
\end{itemize}

Hence, there exist $3 \times 3$ different setups for two premises. It is a refinement of the experimental setup used by~\citet{NEURIPS2020_e992111e}, which only distinguishes whether a premise is explicitly included in the input. We determine the memorization of a premise by the probing results of the memorizing task, which will be elaborated in Sec.~\ref{sec:split}.
 
\section{Probing Methods}
\label{sec:methods}

We investigate encoder-based PLMs (BERT~\cite{devlin-etal-2019-bert} and RoBERTa~\cite{Liu2019RoBERTaAR}) that can be utilized as input encoders for various NLP tasks. Prompt is an intuitive method of our probing task as it matches the mask-filling nature of BERT. We use OpenPrompt~\cite{ding-etal-2022-openprompt}, an open-source framework for prompt learning that includes the mainstream prompt methods, to facilitate the experiments.

\subsection{Probing Methods for Memorization}
\subsubsection{Prompt Templates}
\label{sec:template}
\paragraph{Manual Templates} Manual prompts with human-designed templates written in discrete language phrases are widely used in zero-shot probing~\cite{schick-schutze-2021-exploiting} as PLMs can perform tasks without any training. Manual templates are designed for all the ontological relationships in our task, as shown in Table~\ref{template}.

\paragraph{Soft Templates} One of the disadvantages of manual prompts is that the performance can be significantly affected by perturbation to the prompt templates~\cite{jiang-etal-2020-know}. A common alternative is to use soft prompts that consist of learnable soft tokens~\cite{liu2021gpt, li-liang-2021-prefix} instead of manually defined templates. The soft prompts we use for ontological relationships are also shown in Table~\ref{template}. To probe using soft prompts, we tune randomly initialized soft tokens on the training set with the PLMs parameters being frozen. Detailed training setups are listed in Appendix~\ref{appendix:setup}.

\begin{table*}[ht]
\centering
\scalebox{0.75}{
\begin{tabular}{ccc}
\toprule[1.5pt]
\textbf{Ontological Rel.} & \textbf{Manual Template} & \textbf{Soft Template}\\
\midrule
type &  \makecell[l]{Lionel Messi \textbf{\textcolor{purple}{is a}} [MASK] .\\Lionel Messi \textbf{\textcolor{purple}{has class}} [MASK] .\\Lionel Messi \textbf{\textcolor{purple}{is a particular}} [MASK].} &  Lionel Messi \textbf{\textcolor{teal}{<s1> <s2> <s3>}} [MASK] . \\
\midrule
subclass of  &  \makecell[l]{Person \textbf{\textcolor{purple}{is a}} [MASK] .\\Person \textbf{\textcolor{purple}{has superclass}} [MASK] .\\Person \textbf{\textcolor{purple}{is a particular}} [MASK].} &  Person \textbf{\textcolor{teal}{<s1> <s2> <s3>}} [MASK] . \\
\midrule
subproperty of &  Member of sports team \textbf{\textcolor{purple}{implies}} [MASK] . &  Member of sports team \textbf{\textcolor{teal}{<s1> <s2> <s3>}} [MASK] . \\
\midrule
domain &  \makecell[l]{\textbf{\textcolor{purple}{One has to be a particular}} [MASK]\\ \textbf{\textcolor{purple}{to}} be a player at a sports team .} &  Member of sports team \textbf{\textcolor{teal}{<s1> <s2> <s3>}} [MASK] . \\
\midrule
range &  \makecell[l]{\textbf{\textcolor{purple}{One has to be a particular}} [MASK]\\ \textbf{\textcolor{purple}{to}} have a player at that .} &  Member of sports team \textbf{\textcolor{teal}{<s1> <s2> <s3>}} [MASK] . \\
\bottomrule[1.5pt]
\end{tabular}}
\caption{Manual and soft templates used in prompt-based probing. In soft templates, <s1> <s2> and <s3> correspond to soft tokens.}
\label{template}
\end{table*}

\subsubsection{Candidates Scoring}
\label{sec:candidate_scoring}
 Given a candidate $c$ which can be tokenized into $n$ tokens $c_1, c_2, \dots, c_n$, such that $c_i \in V, i = \{1,\dots, n\}, n \geq 1$, where $V$ is the vocabulary of the model, it is scored based on the log probability of predicting it in the masked prompt. We can either use $n$ different [MASK] tokens or the same [MASK] token to obtain the log probability of each composing token $c_i$, and then compute the log probability of the candidate $c$. For simplicity, we use a single [MASK] token when illustrating our prompts. 

\paragraph{Multiple Masks} For a candidate $c$ consisting of $n$ tokens, we use $n$ [MASK] tokens in the masked input, with the $i$th [MASK] token denoted as $[MASK]_i$. The candidate probability can be computed by three different pooling methods: (1) \textit{mean}: the average of log probabilities of composing tokens~\cite{klein-nabi-2020-contrastive}, (2) \textit{max}: the maximum log probability of all composing tokens, (3) \textit{first}: the log probability of the first composing token. Formally, the score $s$ of candidate $c$ is computed as:
$$\begin{aligned}
\hat{s}_i &= \log{(p([MASK]_i=c_i))} \\
s &= \text{Pooling}(\hat{s}_1,\hat{s}_2, \dots, \hat{s}_n)
\end{aligned}$$

\paragraph{Single Mask} We use one single [MASK] token to obtain an independent prediction of each token. The log probability of each composing token $c_i$ equals the log probability of recovering $c_i$ in the same [MASK], and the candidate is scored with the proposed pooling methods.
\begin{equation}
   \hat{s}_i = \log{(p([MASK]=c_i))} \nonumber
\end{equation}

\subsubsection{Metrics}
\label{sec:metric}
We rank the candidates by their log probability scores and use the top K Recall (R@K) and Mean Reciprocal Rank (MRR) as our evaluation metrics. Since MRR only evaluates the ability to retrieve the first ground truth, we additionally take the average rank of all gold labels as the final rank when computing mean reciprocal rank to evaluate models' ability to retrieve all the ground truths and denote it as MRR$_{a}$. Formally, MRR$_{a}$ is defined as:
\begin{equation}
    \textrm{MRR}_{a} = \frac{1}{n} \sum_{i=1}^n 1/(\frac{1}{|G_i|}\sum_{g \in G_i} \textrm{rank}(g)) \nonumber
\end{equation}
where $n$ is the number of samples in the dataset and $G_i$ is the gold label set of the $i$th sample.

\subsection{Probing Methods for Reasoning}

We explain how we concatenate the premises and hypothesis in the textual input, exclude the models' memory of hypotheses and split a set of premises based on how well the knowledge they represent is memorized by the model. We follow the candidate scoring methods proposed in Sec.~\ref{sec:candidate_scoring} and evaluation metrics in Sec.~\ref{sec:metric}.

\subsubsection{Prompt Templates}
Apart from the prompt templates for our concerned ontological relationships introduced in Sec.~\ref{sec:template}, we further add conjunction tokens between the premises and hypothesis, which can be either manually designed or automatically tuned.

\paragraph{Manual Conj.} As in Figure~\ref{f1}(b), we use a conjunctive adverb \textit{therefore} between the premises and hypothesis. It is kept when there is no premise explicitly given in the input to exclude the effect of the template on probing results under different premise settings.

\paragraph{Soft Conj.} We can also use soft conjunctions by adding a soft token between premises explicitly given in the input and a soft token between the premises and the hypothesis. Therefore, the input would be "$\mathcal{P}_1$ \textbf{\textcolor{teal}{<s4>}} $\mathcal{P}_2$ \textbf{\textcolor{teal}{<s5>}} $\mathcal{H}$". The soft templates used in $\mathcal{P}_1, \mathcal{P}_2$ and $\mathcal{H}$ are loaded from the learned soft prompts in memorizing tasks and finetuned together with soft conjunctions.

\subsubsection{Reasoning with Pseudowords}
\label{sec:pw}

When testing the reasoning ability of PLMs, we replace the specific instances, classes, and properties in the hypothesis prompt with \textit{pseudowords} to prevent probing the memorization of hypotheses. Pseudowords~\cite{schutze-1998-automatic, Zhang2022WordKD, goodwin-etal-2020-probing} are artificially constructed words without any specific lexical meaning. For example, the reasoning prompt for the transitivity of subclass (i.e., rule rdfs9) is "[X] is a person. Person is an animal. Therefore, [X] is a particular [MASK] .", where [X] is a pseudoword. 

Inspired by~\citep{karidi-etal-2021-putting}, we obtain pseudowords for PLMs by creating embeddings without special semantics. Specifically, we sample embeddings at a given distance from the [MASK] token, as the [MASK] token can be used to predict all the words in the vocabulary and appear anywhere in the sentence. The sampling distance $d$ is set to be smaller than the minimum L2 distance between [MASK] and any other tokens in the static embedding space. Formally:
\begin{equation}
    d = \alpha \cdot \min_{t \in V} \Vert \mathbf{z}_t - \mathbf{z}_{[MASK]} \Vert_2 \nonumber
\end{equation}
where $\mathbf{z}_t$ is the static embedding of token $t$ and $\alpha \in (0,1)$ is a coefficient. Moreover, we require that the distance between two pseudowords is at least the sampling distance $d$ to ensure they can be distinguished from each other. 

\subsubsection{Classifying Premises: Memorized or not}
\label{sec:split}

To determine whether a premise is memorized by the model when it is not explicitly given in the input, we employ a classifying method based on the rank of the correct answer in the memorizing task to sort and divide the premise set. The first half of the premise set is regarded as memorized, and the second half is not.

Each rule consists of two premises and we classify them separately. For $\mathcal{P}_1$, which involves knowledge of subclass, subproperty, domain or range tested in the memorizing task, we can leverage previously calculated reciprocal rank during the evaluation. Premises are then sorted in descending order by the reciprocal rank. We conduct the same tests on $\mathcal{P}_2$, which involves knowledge of pseudowords, to examine model predispositions towards specific predictions and classify whether $\mathcal{P}_2$ is memorized or not. Finally, we form our test set by combining premises according to the entailment rule and how each premise is given.

\section{Results and Findings}
In this section, we introduce the performance of PLMs\footnote{We use variants of BERT and RoBERTa models from \url{https://huggingface.co}.} on the test sets of memorizing and reasoning tasks, and analyze the results to posit a series of findings. We then analyze the effectiveness of different prompts. Detailed experimental results can be found in Appendix~\ref{sec:appendix}.

\subsection{Memorizing Task}
\label{sec:mem}
\begin{table*}[ht]
\centering
\scalebox{0.7}{
\begin{tabular}{c|c|ccccccccccccc}
\toprule[1.5pt]
\multirow{3}{*}{Task} &
  \multirow{3}{*}{Metric} &
  \multicolumn{13}{c}{Model} \\ \cmidrule{3-15} 
 &
   &
  \multicolumn{1}{c|}{\multirow{2}{*}{\begin{tabular}[c]{@{}c@{}}Frequency   \\      Baseline\end{tabular}}} &
  \multicolumn{2}{c|}{BERT-B-C} &
  \multicolumn{2}{c|}{BERT-B-U} &
  \multicolumn{2}{c|}{BERT-L-C} &
  \multicolumn{2}{c|}{BERT-L-U} &
  \multicolumn{2}{c|}{RoBERTa-B} &
  \multicolumn{2}{c}{RoBERTa-L} \\
 &
   &
  \multicolumn{1}{c|}{} &
  manT &
  \multicolumn{1}{c|}{softT} &
  manT &
  \multicolumn{1}{c|}{softT} &
  manT &
  \multicolumn{1}{c|}{softT} &
  manT &
  \multicolumn{1}{c|}{softT} &
  manT &
  \multicolumn{1}{c|}{softT} &
  manT &
  softT \\ \midrule
\multirow{4}{*}{TP} &
  R@1 &
  \multicolumn{1}{c|}{15.4} &
  18.9 &
  \multicolumn{1}{c|}{20.1} &
  21.2 &
  \multicolumn{1}{c|}{\textbf{24.8}} &
  15.7 &
  \multicolumn{1}{c|}{22.9} &
  22.3 &
  \multicolumn{1}{c|}{13.1} &
  6.6 &
  \multicolumn{1}{c|}{15.9} &
  9.0 &
  8.7 \\
 &
  R@5 &
  \multicolumn{1}{c|}{15.6} &
  41.0 &
  \multicolumn{1}{c|}{46.4} &
  48.8 &
  \multicolumn{1}{c|}{49.3} &
  46.3 &
  \multicolumn{1}{c|}{\textbf{50.6}} &
  42.1 &
  \multicolumn{1}{c|}{43.9} &
  18.3 &
  \multicolumn{1}{c|}{41.1} &
  39.1 &
  22.4 \\
 &
  MRR$_a$ &
  \multicolumn{1}{c|}{1.3} &
  2.0 &
  \multicolumn{1}{c|}{1.9} &
  \textbf{3.1} &
  \multicolumn{1}{c|}{2.7} &
  2.4 &
  \multicolumn{1}{c|}{2.0} &
  1.8 &
  \multicolumn{1}{c|}{2.0} &
  0.9 &
  \multicolumn{1}{c|}{1.9} &
  1.6 &
  0.9 \\
 &
  MRR &
  \multicolumn{1}{c|}{19.6} &
  28.4 &
  \multicolumn{1}{c|}{31.2} &
  33.2 &
  \multicolumn{1}{c|}{35.1} &
  25.0 &
  \multicolumn{1}{c|}{\textbf{36.0}} &
  32.1 &
  \multicolumn{1}{c|}{23.9} &
  11.9 &
  \multicolumn{1}{c|}{28.1} &
  23.7 &
  14.9 \\ \midrule
\multirow{4}{*}{SCO} &
  R@1 &
  \multicolumn{1}{c|}{8.1} &
  11.0 &
  \multicolumn{1}{c|}{29.7} &
  15.1 &
  \multicolumn{1}{c|}{\textbf{37.9}} &
  14.0 &
  \multicolumn{1}{c|}{35.0} &
  11.6 &
  \multicolumn{1}{c|}{31.0} &
  9.8 &
  \multicolumn{1}{c|}{24.5} &
  9.0 &
  22.8 \\
 &
  R@5 &
  \multicolumn{1}{c|}{38.9} &
  38.1 &
  \multicolumn{1}{c|}{47.9} &
  43.5 &
  \multicolumn{1}{c|}{\textbf{55.9}} &
  43.8 &
  \multicolumn{1}{c|}{54.6} &
  35.4 &
  \multicolumn{1}{c|}{53.5} &
  22.1 &
  \multicolumn{1}{c|}{41.4} &
  39.1 &
  42.8 \\
 &
  MRR$_a$ &
  \multicolumn{1}{c|}{7.4} &
  5.3 &
  \multicolumn{1}{c|}{11.8} &
  6.6 &
  \multicolumn{1}{c|}{\textbf{13.3}} &
  6.7 &
  \multicolumn{1}{c|}{9.7} &
  3.7 &
  \multicolumn{1}{c|}{8.9} &
  4.2 &
  \multicolumn{1}{c|}{8.5} &
  4.5 &
  5.5 \\
 &
  MRR &
  \multicolumn{1}{c|}{23.7} &
  22.7 &
  \multicolumn{1}{c|}{39.2} &
  29.0 &
  \multicolumn{1}{c|}{\textbf{46.4}} &
  25.8 &
  \multicolumn{1}{c|}{41.2} &
  21.9 &
  \multicolumn{1}{c|}{41.9} &
  16.7 &
  \multicolumn{1}{c|}{29.7} &
  24.6 &
  32.9 \\ \midrule
\multirow{4}{*}{SPO} &
  R@1 &
  \multicolumn{1}{c|}{25.6} &
  23.1 &
  \multicolumn{1}{c|}{38.5} &
  20.5 &
  \multicolumn{1}{c|}{38.5} &
  18.0 &
  \multicolumn{1}{c|}{38.5} &
  23.1 &
  \multicolumn{1}{c|}{\textbf{41.0}} &
  10.3 &
  \multicolumn{1}{c|}{35.9} &
  10.3 &
  41.0 \\
 &
  R@5 &
  \multicolumn{1}{c|}{28.2} &
  64.1 &
  \multicolumn{1}{c|}{64.1} &
  69.2 &
  \multicolumn{1}{c|}{\textbf{74.4}} &
  59.0 &
  \multicolumn{1}{c|}{76.9} &
  69.2 &
  \multicolumn{1}{c|}{64.1} &
  33.3 &
  \multicolumn{1}{c|}{61.5} &
  30.8 &
  69.2 \\
 &
  MRR$_a$ &
  \multicolumn{1}{c|}{15.8} &
  15.8 &
  \multicolumn{1}{c|}{23.8} &
  19.5 &
  \multicolumn{1}{c|}{29.3} &
  19.5 &
  \multicolumn{1}{c|}{\textbf{29.8}} &
  19.0 &
  \multicolumn{1}{c|}{28.8} &
  8.8 &
  \multicolumn{1}{c|}{25.1} &
  10.0 &
  29.6 \\
 &
  MRR &
  \multicolumn{1}{c|}{31.2} &
  39.2 &
  \multicolumn{1}{c|}{43.7} &
  38.3 &
  \multicolumn{1}{c|}{53.5} &
  34.5 &
  \multicolumn{1}{c|}{49.8} &
  39.3 &
  \multicolumn{1}{c|}{52.9} &
  20.6 &
  \multicolumn{1}{c|}{47.4} &
  21.9 &
  \textbf{53.8} \\ \midrule
\multirow{3}{*}{DM} &
  R@1 &
  \multicolumn{1}{c|}{43.3} &
  43.3 &
  \multicolumn{1}{c|}{30.0} &
  43.3 &
  \multicolumn{1}{c|}{40.0} &
  \textbf{50.0} &
  \multicolumn{1}{c|}{40.0} &
  33.3 &
  \multicolumn{1}{c|}{26.7} &
  6.7 &
  \multicolumn{1}{c|}{43.3} &
  13.3 &
  16.7 \\
 &
  R@5 &
  \multicolumn{1}{c|}{60.0} &
  53.3 &
  \multicolumn{1}{c|}{60.0} &
  53.3 &
  \multicolumn{1}{c|}{\textbf{63.3}} &
  60.0 &
  \multicolumn{1}{c|}{\textbf{63.3}} &
  53.3 &
  \multicolumn{1}{c|}{50.0} &
  20.0 &
  \multicolumn{1}{c|}{\textbf{63.3}} &
  46.7 &
  50.0 \\
 &
  MRR &
  \multicolumn{1}{c|}{\textbf{50.9}} &
  47.6 &
  \multicolumn{1}{c|}{40.7} &
  49.3 &
  \multicolumn{1}{c|}{50.0} &
  50.3 &
  \multicolumn{1}{c|}{48.7} &
  43.2 &
  \multicolumn{1}{c|}{33.5} &
  15.3 &
  \multicolumn{1}{c|}{49.0} &
  27.4 &
  25.5 \\ \midrule
\multirow{3}{*}{RG} &
  R@1 &
  \multicolumn{1}{c|}{10.7} &
  46.4 &
  \multicolumn{1}{c|}{\textbf{57.1}} &
  42.9 &
  \multicolumn{1}{c|}{\textbf{57.1}} &
  \textbf{57.1} &
  \multicolumn{1}{c|}{\textbf{57.1}} &
  46.4 &
  \multicolumn{1}{c|}{53.6} &
  32.1 &
  \multicolumn{1}{c|}{46.4} &
  17.9 &
  42.9 \\
 &
  R@5 &
  \multicolumn{1}{c|}{53.6} &
  67.9 &
  \multicolumn{1}{c|}{67.9} &
  75.0 &
  \multicolumn{1}{c|}{75.0} &
  \textbf{78.6} &
  \multicolumn{1}{c|}{75.0} &
  78.6 &
  \multicolumn{1}{c|}{75.0} &
  57.1 &
  \multicolumn{1}{c|}{53.6} &
  53.6 &
  71.4 \\
 &
  MRR &
  \multicolumn{1}{c|}{31.2} &
  59.1 &
  \multicolumn{1}{c|}{62.7} &
  56.0 &
  \multicolumn{1}{c|}{63.9} &
  \textbf{66.8} &
  \multicolumn{1}{c|}{66.2} &
  61.1 &
  \multicolumn{1}{c|}{59.5} &
  44.0 &
  \multicolumn{1}{c|}{50.3} &
  33.2 &
  48.5\\
  \bottomrule[1.5pt]
\end{tabular}}
\caption{Performance (\%) of the memorizing task. B/L stands for base/large and C/U stands for cased/uncased. The distinction between the prompt templates (manT for manual template and softT for soft template) is preserved, and for the other settings, such as the number of [MASK] tokens and pooling methods, we use the ones that give the best results and discuss their impacts in Appendix~\ref{appendix:prompt}.}
\label{mem_results}
\end{table*}

The baseline model used for the memorizing task is a frequency-based model which predicts a list of gold labels in the training set based on the frequency at which they appear, followed by a random list of candidates that are not gold labels in the training set. It combines prior knowledge and random guesses and is stronger than a random baseline. 

The experimental results of the memorizing task are summarized in Table~\ref{mem_results}, from which we can observe that: (1) The best performance of PLMs is better than the baseline on every task except for DM. On DM, the baseline achieves higher MRR. If taking all three metrics into account, the best performance of PLMs still surpasses the performance of the baseline. (2) Except for DM, BERT models achieve much better performance than the baseline in all subtasks and all metrics. Taking an average of the increase in each metric, they outperform the baseline by 43--198\%. Only BERT-base-uncased and BERT-large-cased outperform the baseline in DM by a small margin of 1\% and 7\%. (3) RoBERTa models generally fall behind BERT, showing a 38--134\% improvement compared with the baseline except for DM. (4) Despite a significant improvement from the baseline, the results are still not perfect in all subtasks.

\paragraph{PLMs can memorize certain ontological knowledge but not perfectly.} Based on the above observation, we can conclude that PLMs have a certain memory of the concerned ontological relationships and the knowledge can be accessed via prompt, allowing them to outperform a strong baseline. It proves that during pretraining, language models learn not only facts about entities but also their ontological relationships, which is essential for a better organization of world knowledge. However, the memorization is not perfect, urging further efforts on ontology-aware pretraining.

\paragraph{Large models are not necessarily better at memorizing ontological knowledge.} According to~\citet{petroni-etal-2019-language}, models with larger sizes appear to store more knowledge and achieve better performance in both knowledge probing tasks and downstream NLP tasks. However, as shown in Table~\ref{mem_results}, BERT-large-uncased is worse than its smaller variant under most circumstances, and RoBERTa-large is worse than RoBERTa-base in TP and DM. It demonstrates that the scale of model parameters does not necessarily determine the storage of ontological knowledge.

\subsection{Reasoning Task}
\label{sec:reasoning}
\begin{figure*}[ht]
    \centering
    \includegraphics[width=0.95\textwidth]{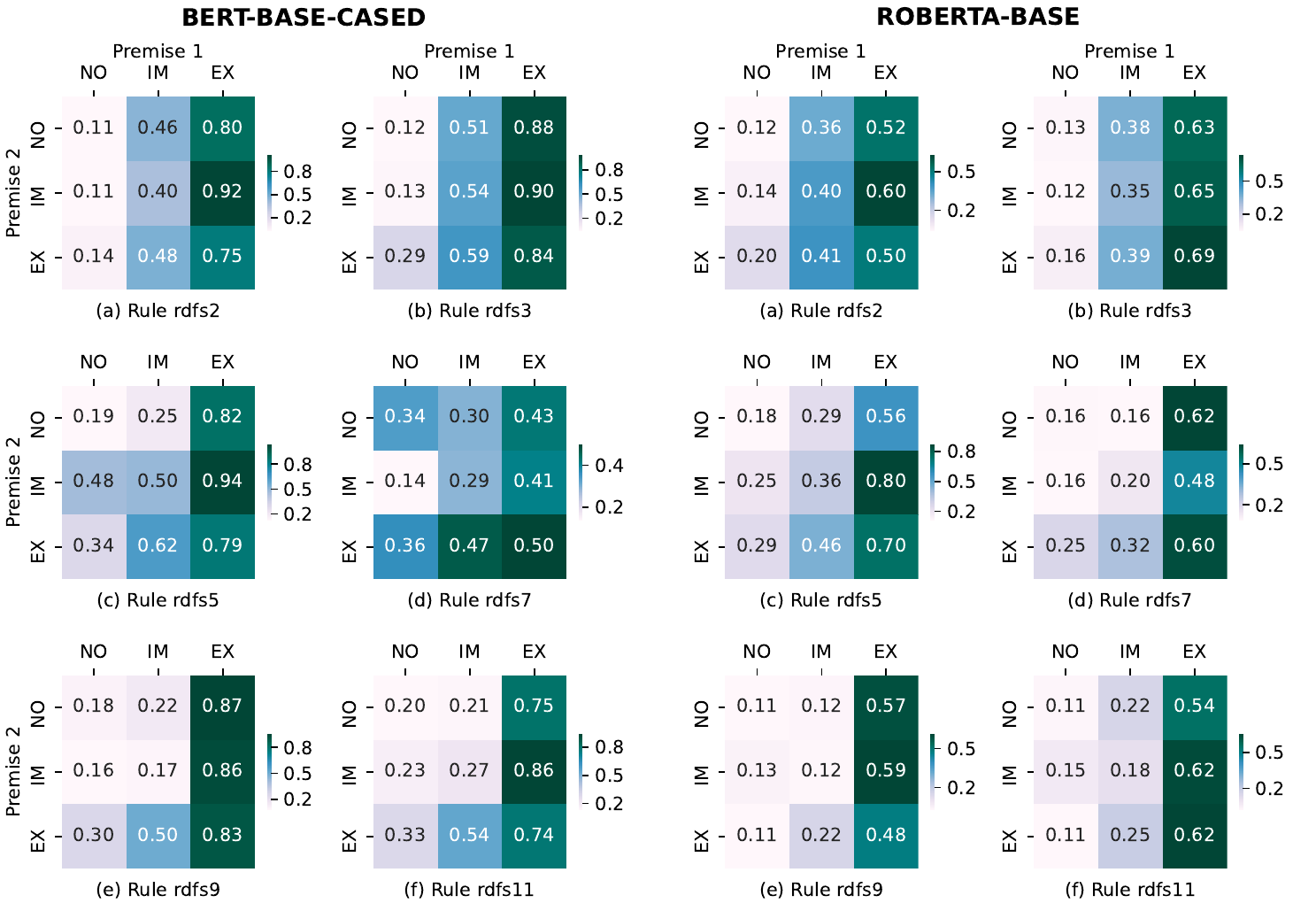}
    \caption{The MRR by BERT-base-cased and RoBERTa-base using different combinations of premises. EX stands for explicitly given, IM stands for implicitly given and NO stands for not given. Other metrics show similar trends.}
    \label{reasoning}
\end{figure*}

We fix the usage of multiple masks and mean-pooling in the reasoning experiments as they generally outperform other settings in the memorizing task (see Appendix~\ref{appendix:prompt}). We take an average of the MRR metrics using different templates and illustrate the results of BERT-base-cased and RoBERTa-base in Figure~\ref{reasoning}. With neither premise given, the rank of the ground truth is usually low. It shows that models have little idea of the hypothesis, which is reasonable because the information of pseudowords is probed. With premises implicitly or explicitly given, especially $\mathcal{P}_1$, the MRR metrics improve in varying degrees. Moreover, results show that BERT-base-cased has better reasoning ability with our concerned ontological entailment rules than RoBERTa-base.

\begin{figure}[htbp]
    \centering
    \includegraphics[width=0.3\textwidth]{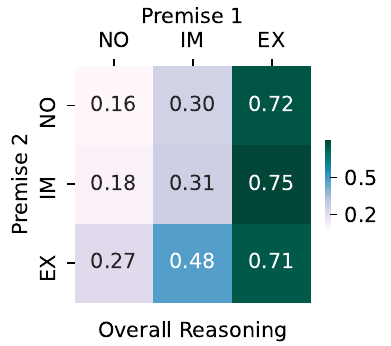}
    \caption{The macro-averaged MRR across different entailment rules and language models with different combinations of premises.}
    \label{overall_reasoning}
\end{figure}

\paragraph{PLMs have a limited understanding of the semantics behind ontological knowledge.} To reach a more general conclusion, we illustrate the overall reasoning performance in Figure~\ref{overall_reasoning} by averaging over all the entailment rules and PLMs, and find that: (1) When $\mathcal{P}_1$ is explicitly given in the input text, models are able to significantly improve the rank of gold labels. As $\mathcal{P}_1$ contains the ground truth in its context, it raises doubt about whether the improvement is obtained through logical reasoning or just priming~\cite{misra-etal-2020-exploring}. 
(2) Explicitly giving $\mathcal{P}_2$ introduces additional tokens that may not be present in gold labels, making $\mathcal{P}_1/\mathcal{P}_2=\text{EX}/\text{EX}$ worse than $\mathcal{P}_1/\mathcal{P}_2=\text{EX}/\text{IM}$. (3) When premises are implicitly given, the MRR metrics are higher than when they are not given. It implies that, to some extent, PLMs can utilize the implicit ontological knowledge and select the correct entailment rule to make inferences. (4) However, none of the premises combinations can give near-perfect reasoning performance (MRR metrics close to 1), suggesting that PLMs only have a weak understanding of ontological knowledge.


\paragraph{Paraphrased properties are a challenge for language models.} In Figure~\ref{reasoning}(d), the premise $\mathcal{P}_1$ of rule rdfs7 contains a paraphrased version of the ground truth, which is the manually-designed pattern of a particular property. Compared with rule rdfs5 shown in Figure~\ref{reasoning}(c), where $\mathcal{P}_1$ contains the surface form of the correct property, the MRR of BERT-base-cased of rdfs7 decreases by 23\%, 49\% and 29\% when $\mathcal{P}_1$ is explicitly given and $\mathcal{P}_2$ is not, implicitly and explicitly given, respectively. Though the MRR of RoBERTa-base of rdfs7 increases when $\mathcal{P}_2$ is not given, it decreases by 40\% and 15\% when $\mathcal{P}_2$ is implicitly and explicitly given. This suggests that PLMs fail to understand the semantics of some properties, thus demonstrating a limited understanding of ontological knowledge.

\subsection{Effectiveness of Prompts} 
In this section, we discuss how prompt templates affect performance. In the memorizing task, Table~\ref{mem_results} shows that using soft templates generally improves the performance of memorizing tasks, in particular TP, SCO and SPO. It suggests that it is non-trivial to extract knowledge from PLMs. 

Meanwhile, only a few models perform better with soft templates on DM and RG with a relatively marginal improvement. This could be explained by the fact that both the manual templates and semantics of domain and range constraints are more complex than those of other relationships. Therefore, it is difficult for models to capture with only three soft tokens. We also note that RoBERTa models appear to benefit more from soft templates than BERT models, probably due to their poor performance with manual templates.

\begin{figure}[htbp]
    \centering
    \includegraphics[width=0.45\textwidth]{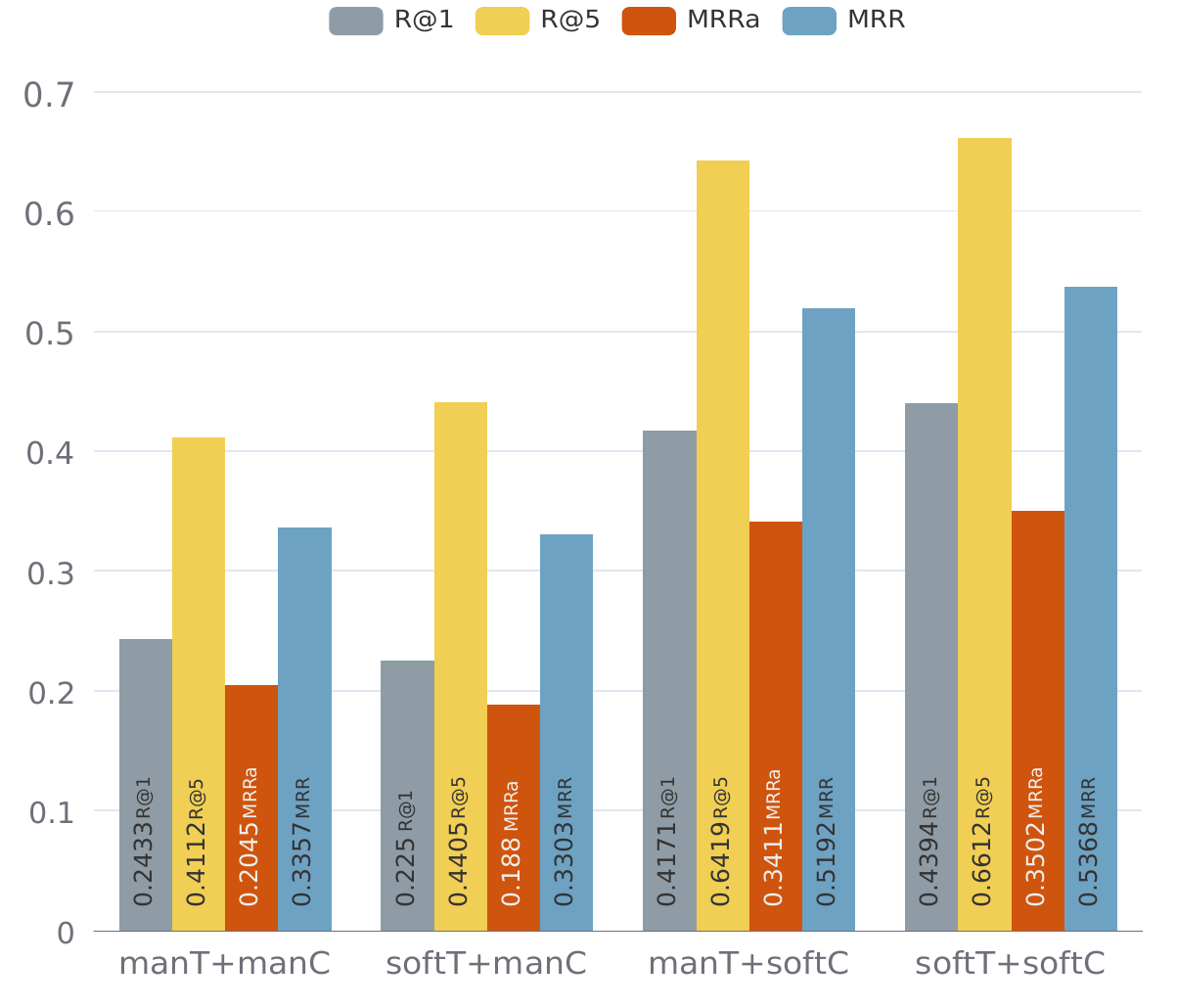}
    \caption{Effectiveness of different combinations of templates and conjunction tokens in reasoning.}
    \label{prompts}
\end{figure}

Trained soft templates for each relation barely help with reasoning, though. In Figure~\ref{prompts}, we summarize the performance by averaging across different models and reasoning tasks and find that it is the trained conjunction token which improves the performance of reasoning rather than the soft templates that describe ontological relationships. It might be inspiring that natural language inference with PLMs can be improved by adding trainable tokens as conjunctions instead of simply concatenating all the premises.

\section{Preliminary Evaluation of ChatGPT}

After we finished the majority of our probing experiments, ChatGPT, a decoder-only model, was publicly released and demonstrated remarkable capabilities in commonsense knowledge and reasoning. Therefore, we additionally perform a preliminary probe of the ability of ChatGPT to memorize and understand ontological knowledge.

Since ChatGPT is a decoder-only model, we employ a distinct probing method from what is expounded in Sec.~\ref{sec:methods}. Instead of filling masks, we directly ask ChatGPT to answer multiple-choice questions with 20 candidate choices and evaluate the accuracy.

\subsection{Probing for Memorization Ability}\label{sec:gpt-mem}

For memorization probing, we use the finest-grained gold label as the correct answer and randomly sample 19 negative candidates to form the choice set. Take the TP task as an example, we query the GPT-3.5-turbo API with the prompt "What is the type of Lionel Messi? (a) soccer player, (b) work, (c) ..." followed by remaining candidates. We sample 500 test cases for the TP and SCO tasks and use the complete test sets for the other tasks.


\begin{table}[tb]
\centering
\tabcolsep=0.15cm
\scalebox{0.8}{
\begin{tabular}{ccc}
\toprule[1.5pt]
 Task & ChatGPT & BERT-base-uncased\\
\midrule
TP & \textbf{70.2} & 42.6\\
SCO &  \textbf{83.6} & 52.4\\
SPO & \textbf{71.8}  & 38.5\\
DM & \textbf{86.7} & 70.0\\
RG & \textbf{82.1} & \textbf{82.1} \\
\bottomrule[1.5pt]
\end{tabular}}
\caption{Accuracy (\%) achieved by ChatGPT and BERT-base-uncased on the multiple-choice memorizing task with 20 candidates.}
\label{gpt-mem}
\end{table}

For comparison, we also conduct the experiments using BERT-base-uncased, a generally competitive PLM in memorizing and understanding ontological knowledge, with manual prompts and the identical candidate subset. The results presented in Table~\ref{gpt-mem} indicate that ChatGPT outperforms BERT-base-uncased significantly in most of the memorizing tasks associated with ontological knowledge. 

\subsection{Probing for Reasoning Ability}

Since we cannot input embeddings in the GPT-3.5-turbo API, we use \textit{X} and \textit{Y} to represent pseudowords as they are single letters that do not convey meanings. However, ChatGPT cannot generate any valid prediction without sufficient context regarding these pseudowords. Therefore, $\mathcal{P}_2$ needs to be explicitly provided to describe the characteristics or relations of the pseudowords. We then explore the ability of ChatGPT to select the correct answer from 20 candidates with different forms of $\mathcal{P}_1$. In this task, $\mathcal{P}_1$ is regarded as memorized if the model can correctly choose the gold answer from the given 20 candidates in the memorizing task.



\begin{table}[tb]
\centering
\scalebox{0.75}{
\begin{tabular}{c|c|cccccc}
\toprule[1.5pt]
\multirow{2}{*}{$\mathcal{P}_1$} & \multirow{2}{*}{AVG} &  \multicolumn{6}{c}{RDFS Rule}                  \\ \cmidrule{3-8} 
         &          & rdfs2 & rdfs3 & rdfs5 & rdfs7 & rdfs9 & rdfs11 \\ \midrule
NO          &  13.5    & 25.0 & 16.7 & 0.0 & 0.0 & 19.0 & 20.8\\
IM          & 82.8      & 76.9 & 86.4 & 71.5 &  77.7 &  91.9 & 92.4 \\
EX           & 97.1       & 100.0  & 96.4 & 94.9 & 96.9 & 97.4 & 97.0 \\
\bottomrule[1.5pt]
\end{tabular}
}
\caption{Accuracy (\%) achieved by ChatGPT on each reasoning subtask with $\mathcal{P}_2$ explicitly given.}
\label{gpt-rea}
\end{table}

Based on the results presented in Table~\ref{gpt-rea}, ChatGPT demonstrates high accuracy when $\mathcal{P}_1$ is either implicitly or explicitly given, suggesting its strong capacity to reason and understand ontological knowledge. 
Due to a substantial disparity in the knowledge memorized by ChatGPT compared to other models (as shown in section \ref{sec:gpt-mem}), 
their performance is not directly comparable when $\mathcal{P}_1$ is not given or implicitly given. 
Therefore, we only compare ChatGPT and BERT-base-uncased when $\mathcal{P}_1$ is explicitly given. Results show that ChatGPT significantly outperforms BERT-base-uncased in explicit reasoning (97.1\% vs. 88.2\%).


\section{Related Work}
\paragraph{Knowledge Probing}
Language models are shown to encode a wide variety of knowledge after being pretrained on a large-scale corpus. Recent studies probe PLMs for linguistic knowledge~\cite{vulic-etal-2020-probing, hewitt-manning-2019-structural}, world knowledge~\cite{petroni-etal-2019-language, jiang-etal-2020-know, safavi-koutra-2021-relational}, actionable knowledge~\cite{Huang2022LanguageMA}, etc. via methods such as cloze prompts~\cite{beloucif-biemann-2021-probing-pre, petroni-2020-context} and linear classifiers~\cite{hewitt-liang-2019-designing, pimentel-etal-2020-information}. Although having explored extensive knowledge within PLMs, previous knowledge probing works have not studied ontological knowledge systematically. We cut through this gap to investigate how well PLMs know about ontological knowledge and the meaning behind the surface form.

\paragraph{Knowledge Reasoning}
Reasoning is the process of drawing new conclusions through the use of existing knowledge and rules. Progress has been reported in using PLMs to perform reasoning tasks, including arithmetic~\cite{Wang2022SelfConsistencyIC, Wei2022ChainOT}, commonsense~\cite{talmor-etal-2019-commonsenseqa, NEURIPS2020_e992111e, Wei2022ChainOT}, logical~\cite{Creswell2022SelectionInferenceEL} and symbolic reasoning~\cite{Wei2022ChainOT}. These abilities can be unlocked by finetuning a classifier on downstream datasets~\cite{NEURIPS2020_e992111e} or using proper prompting strategies (e.g., chain of thought (CoT) prompting~\cite{Wei2022ChainOT} and generated knowledge prompting~\cite{liu-etal-2022-generated}). This suggests that despite their insensitivity to negation~\cite{ettinger-2020-bert, kassner-schutze-2020-negated} and over-sensitivity to lexicon cues like priming words~\cite{helwe2021reasoning, misra-etal-2020-exploring}, PLMs have the potential to make inferences over implicit knowledge and explicit natural language statements. In this work, we investigate the ability of PLMs to perform logical reasoning with implicit ontological knowledge to examine whether they understand the semantics beyond memorization.
\section{Conclusion}
In this work, we systematically probe whether PLMs encode ontological knowledge and understand its semantics beyond the surface form. Experiments show that PLMs can memorize some ontological knowledge and make inferences based on implicit knowledge following ontological entailment rules, suggesting that PLMs possess a certain level of awareness and understanding of ontological knowledge. However, it is important to note that both the accuracy of memorizing and reasoning is less than perfect, and the difficulty encountered by PLMs when processing paraphrased knowledge is confirmed. These observations indicate that their knowledge and understanding of ontology are limited. Therefore, enhancing the knowledge and understanding of ontology would be a worthy future research goal for language models. Our exploration into ChatGPT shows an improved performance in both memorizing and reasoning tasks, signifying the potential for further advancements.
\section*{Limitations}
The purpose of our work is to evaluate the ontological knowledge of PLMs. However, a sea of classes and properties exist in the real world and we only cover a selective part of them. Consequently, the scope of our dataset for the experimental analysis is limited.  
The findings from our experiments demonstrate an imperfect knowledge and understanding obtained by the models, indicating a tangible room for enhancement in both ontological knowledge memorization and understanding and a need for a better ability to address paraphrasing. These observations lead us to contemplate refining the existing pretraining methods to help language models achieve better performance in related tasks.

\section*{Ethics Statement}
We propose our ethics statement of the work in this section: 
 (1) Dataset. Our data is obtained from DBpedia and Wikidata, two publicly available linked open data projects related to Wikipedia. Wikidata is under the Creative Commons CC0 License, and DBpedia is licensed under the terms of the Creative Commons Attribution-ShareAlike 3.0 license and the GNU Free Documentation License. We believe the privacy policies of DBpedia\footnote{\url{https://www.dbpedia.org/privacy/}} and Wikidata\footnote{\url{https://foundation.wikimedia.org/wiki/Privacy_policy}} are well carried out. We inspect whether our dataset, especially instances collected, contains any unethical content. No private information or offensive topics are found during human inspection. (2) Labor considerations. During dataset construction, the authors voluntarily undertake works requiring human efforts, including data collection, cleansing, revision and design of property patterns. All the participants are well informed about how the dataset will be processed, used and released. (3) Probing results. As PLMs are pretrained on large corpora, they may give biased results when being probed. We randomly check some probing results and find no unethical content in these samples. Therefore, we believe that our study does not introduce additional risks.

\section*{Acknowledgement}
This work was supported by the National Natural Science Foundation of China (61976139) and by Alibaba Group through Alibaba Innovative Research Program.

\bibliography{main}

\begin{thebibliography}{44}
\expandafter\ifx\csname natexlab\endcsname\relax\def\natexlab#1{#1}\fi

\bibitem[{AlKhamissi et~al.(2022)AlKhamissi, Li, Celikyilmaz, Diab, and Ghazvininejad}]{AlKhamissi2022ARO}
Badr AlKhamissi, Millicent Li, Asli Celikyilmaz, Mona Diab, and Marjan Ghazvininejad. 2022.
\newblock A review on language models as knowledge bases.
\newblock \emph{arXiv preprint arXiv:2204.06031}.

\bibitem[{Auer et~al.(2007)Auer, Bizer, Kobilarov, Lehmann, Cyganiak, and Ives}]{Auer2007DBpediaAN}
Sören Auer, Christian Bizer, Georgi Kobilarov, Jens Lehmann, Richard Cyganiak, and Zachary Ives. 2007.
\newblock \href {https://doi.org/10.1007/978-3-540-76298-0_52} {Dbpedia: A nucleus for a web of open data}.
\newblock \emph{Lecture Notes in Computer Science}, 6:722--735.

\bibitem[{Beloucif and Biemann(2021)}]{beloucif-biemann-2021-probing-pre}
Meriem Beloucif and Chris Biemann. 2021.
\newblock \href {https://doi.org/10.18653/v1/2021.findings-emnlp.218} {Probing pre-trained language models for semantic attributes and their values}.
\newblock In \emph{Findings of the Association for Computational Linguistics: EMNLP 2021}, pages 2554--2559, Punta Cana, Dominican Republic. Association for Computational Linguistics.

\bibitem[{Bhatia and Richie(2020)}]{Bhatia2020TransformerNO}
Sudeep Bhatia and Russell Richie. 2020.
\newblock \href {https://doi.org/10.1037/rev0000319} {Transformer networks of human conceptual knowledge.}
\newblock \emph{Psychological review}.

\bibitem[{Brickley and Guha(2002)}]{Brickley2002ResourceDF}
Dan Brickley and Ramanathan~V. Guha. 2002.
\newblock \href {https://www.immagic.com/eLibrary/ARCHIVES/TECH/W3C/W3C_RDF.pdf} {Resource description framework (rdf) model and syntax specification}.

\bibitem[{Creswell et~al.(2022)Creswell, Shanahan, and Higgins}]{Creswell2022SelectionInferenceEL}
Antonia Creswell, Murray Shanahan, and Irina Higgins. 2022.
\newblock \href {https://doi.org/10.48550/ARXIV.2205.09712} {Selection-inference: Exploiting large language models for interpretable logical reasoning}.
\newblock \emph{ArXiv}, abs/2205.09712.

\bibitem[{Devlin et~al.(2019)Devlin, Chang, Lee, and Toutanova}]{devlin-etal-2019-bert}
Jacob Devlin, Ming-Wei Chang, Kenton Lee, and Kristina Toutanova. 2019.
\newblock \href {https://doi.org/10.18653/v1/N19-1423} {{BERT}: Pre-training of deep bidirectional transformers for language understanding}.
\newblock In \emph{Proceedings of the 2019 Conference of the North {A}merican Chapter of the Association for Computational Linguistics: Human Language Technologies, Volume 1 (Long and Short Papers)}, pages 4171--4186, Minneapolis, Minnesota. Association for Computational Linguistics.

\bibitem[{Ding et~al.(2022)Ding, Hu, Zhao, Chen, Liu, Zheng, and Sun}]{ding-etal-2022-openprompt}
Ning Ding, Shengding Hu, Weilin Zhao, Yulin Chen, Zhiyuan Liu, Haitao Zheng, and Maosong Sun. 2022.
\newblock \href {https://doi.org/10.18653/v1/2022.acl-demo.10} {{O}pen{P}rompt: An open-source framework for prompt-learning}.
\newblock In \emph{Proceedings of the 60th Annual Meeting of the Association for Computational Linguistics: System Demonstrations}, pages 105--113, Dublin, Ireland. Association for Computational Linguistics.

\bibitem[{Ettinger(2020)}]{ettinger-2020-bert}
Allyson Ettinger. 2020.
\newblock \href {https://doi.org/10.1162/tacl_a_00298} {What {BERT} is not: Lessons from a new suite of psycholinguistic diagnostics for language models}.
\newblock \emph{Transactions of the Association for Computational Linguistics}, 8:34--48.

\bibitem[{Gibbins and Shadbolt(2009)}]{Gibbins2009ResourceDF}
Nicholas Gibbins and Nigel Shadbolt. 2009.
\newblock \href {https://eprints.soton.ac.uk/268264/1/gibbins-shadbolt-elis-rdf-v3.pdf} {Resource description framework (rdf)}.

\bibitem[{Goodwin et~al.(2020)Goodwin, Sinha, and O{'}Donnell}]{goodwin-etal-2020-probing}
Emily Goodwin, Koustuv Sinha, and Timothy~J. O{'}Donnell. 2020.
\newblock \href {https://doi.org/10.18653/v1/2020.acl-main.177} {Probing linguistic systematicity}.
\newblock In \emph{Proceedings of the 58th Annual Meeting of the Association for Computational Linguistics}, pages 1958--1969, Online. Association for Computational Linguistics.

\bibitem[{Goodwin and Demner-Fushman(2020)}]{goodwin-demner-fushman-2020-enhancing}
Travis Goodwin and Dina Demner-Fushman. 2020.
\newblock \href {https://doi.org/10.18653/v1/2020.deelio-1.7} {Enhancing question answering by injecting ontological knowledge through regularization}.
\newblock In \emph{Proceedings of Deep Learning Inside Out (DeeLIO): The First Workshop on Knowledge Extraction and Integration for Deep Learning Architectures}, pages 56--63, Online. Association for Computational Linguistics.

\bibitem[{Helwe et~al.(2021)Helwe, Clavel, and Suchanek}]{helwe2021reasoning}
Chadi Helwe, Chlo{\'e} Clavel, and Fabian~M. Suchanek. 2021.
\newblock \href {https://doi.org/10.24432/C5W300} {Reasoning with transformer-based models: Deep learning, but shallow reasoning}.
\newblock In \emph{3rd Conference on Automated Knowledge Base Construction}.

\bibitem[{Hewitt and Liang(2019)}]{hewitt-liang-2019-designing}
John Hewitt and Percy Liang. 2019.
\newblock \href {https://doi.org/10.18653/v1/D19-1275} {Designing and interpreting probes with control tasks}.
\newblock In \emph{Proceedings of the 2019 Conference on Empirical Methods in Natural Language Processing and the 9th International Joint Conference on Natural Language Processing (EMNLP-IJCNLP)}, pages 2733--2743, Hong Kong, China. Association for Computational Linguistics.

\bibitem[{Hewitt and Manning(2019)}]{hewitt-manning-2019-structural}
John Hewitt and Christopher~D. Manning. 2019.
\newblock \href {https://doi.org/10.18653/v1/N19-1419} {{A} structural probe for finding syntax in word representations}.
\newblock In \emph{Proceedings of the 2019 Conference of the North {A}merican Chapter of the Association for Computational Linguistics: Human Language Technologies, Volume 1 (Long and Short Papers)}, pages 4129--4138, Minneapolis, Minnesota. Association for Computational Linguistics.

\bibitem[{hommeaux(2011)}]{hommeaux2011SPARQLQL}
E.~Prud hommeaux. 2011.
\newblock \href {https://www.w3.org/TR/rdf-sparql-query/} {Sparql query language for rdf}.

\bibitem[{Huang et~al.(2022)Huang, Abbeel, Pathak, and Mordatch}]{Huang2022LanguageMA}
Wenlong Huang, Pieter Abbeel, Deepak Pathak, and Igor Mordatch. 2022.
\newblock \href {https://proceedings.mlr.press/v162/huang22a/huang22a.pdf} {Language models as zero-shot planners: Extracting actionable knowledge for embodied agents}.
\newblock In \emph{International Conference on Machine Learning}, pages 9118--9147. PMLR.

\bibitem[{Jiang et~al.(2020)Jiang, Xu, Araki, and Neubig}]{jiang-etal-2020-know}
Zhengbao Jiang, Frank~F. Xu, Jun Araki, and Graham Neubig. 2020.
\newblock \href {https://doi.org/10.1162/tacl_a_00324} {How can we know what language models know?}
\newblock \emph{Transactions of the Association for Computational Linguistics}, 8:423--438.

\bibitem[{Karidi et~al.(2021)Karidi, Zhou, Schneider, Abend, and Srikumar}]{karidi-etal-2021-putting}
Taelin Karidi, Yichu Zhou, Nathan Schneider, Omri Abend, and Vivek Srikumar. 2021.
\newblock \href {https://doi.org/10.18653/v1/2021.emnlp-main.806} {Putting words in {BERT}{'}s mouth: Navigating contextualized vector spaces with pseudowords}.
\newblock In \emph{Proceedings of the 2021 Conference on Empirical Methods in Natural Language Processing}, pages 10300--10313, Online and Punta Cana, Dominican Republic. Association for Computational Linguistics.

\bibitem[{Kassner and Sch{\"u}tze(2020)}]{kassner-schutze-2020-negated}
Nora Kassner and Hinrich Sch{\"u}tze. 2020.
\newblock \href {https://doi.org/10.18653/v1/2020.acl-main.698} {Negated and misprimed probes for pretrained language models: Birds can talk, but cannot fly}.
\newblock In \emph{Proceedings of the 58th Annual Meeting of the Association for Computational Linguistics}, pages 7811--7818, Online. Association for Computational Linguistics.

\bibitem[{Klein and Nabi(2020)}]{klein-nabi-2020-contrastive}
Tassilo Klein and Moin Nabi. 2020.
\newblock \href {https://doi.org/10.18653/v1/2020.acl-main.671} {Contrastive self-supervised learning for commonsense reasoning}.
\newblock In \emph{Proceedings of the 58th Annual Meeting of the Association for Computational Linguistics}, pages 7517--7523, Online. Association for Computational Linguistics.

\bibitem[{Kumar et~al.(2019)Kumar, Kumar, Singh, Patel, and Jain}]{Kumar2019AnOD}
Dikshit Kumar, Agam Kumar, Man Singh, Archana Patel, and Sarika Jain. 2019.
\newblock An online dictionary and thesaurus.

\bibitem[{Li and Liang(2021)}]{li-liang-2021-prefix}
Xiang~Lisa Li and Percy Liang. 2021.
\newblock \href {https://doi.org/10.18653/v1/2021.acl-long.353} {Prefix-tuning: Optimizing continuous prompts for generation}.
\newblock In \emph{Proceedings of the 59th Annual Meeting of the Association for Computational Linguistics and the 11th International Joint Conference on Natural Language Processing (Volume 1: Long Papers)}, pages 4582--4597, Online. Association for Computational Linguistics.

\bibitem[{Lin and Ng(2022)}]{lin-ng-2022-bert}
Ruixi Lin and Hwee~Tou Ng. 2022.
\newblock \href {https://doi.org/10.18653/v1/2022.acl-short.11} {Does {BERT} know that the {IS}-a relation is transitive?}
\newblock In \emph{Proceedings of the 60th Annual Meeting of the Association for Computational Linguistics (Volume 2: Short Papers)}, pages 94--99, Dublin, Ireland. Association for Computational Linguistics.

\bibitem[{Liu et~al.(2022)Liu, Liu, Lu, Welleck, West, Le~Bras, Choi, and Hajishirzi}]{liu-etal-2022-generated}
Jiacheng Liu, Alisa Liu, Ximing Lu, Sean Welleck, Peter West, Ronan Le~Bras, Yejin Choi, and Hannaneh Hajishirzi. 2022.
\newblock \href {https://doi.org/10.18653/v1/2022.acl-long.225} {Generated knowledge prompting for commonsense reasoning}.
\newblock In \emph{Proceedings of the 60th Annual Meeting of the Association for Computational Linguistics (Volume 1: Long Papers)}, pages 3154--3169, Dublin, Ireland. Association for Computational Linguistics.

\bibitem[{Liu et~al.(2021)Liu, Zheng, Du, Ding, Qian, Yang, and Tang}]{liu2021gpt}
Xiao Liu, Yanan Zheng, Zhengxiao Du, Ming Ding, Yujie Qian, Zhilin Yang, and Jie Tang. 2021.
\newblock Gpt understands, too.
\newblock \emph{ArXiv}, abs/2103.10385.

\bibitem[{Liu et~al.(2019)Liu, Ott, Goyal, Du, Joshi, Chen, Levy, Lewis, Zettlemoyer, and Stoyanov}]{Liu2019RoBERTaAR}
Yinhan Liu, Myle Ott, Naman Goyal, Jingfei Du, Mandar Joshi, Danqi Chen, Omer Levy, Mike Lewis, Luke Zettlemoyer, and Veselin Stoyanov. 2019.
\newblock \href {https://doi.org/10.48550/ARXIV.1907.11692} {Roberta: A robustly optimized bert pretraining approach}.
\newblock \emph{ArXiv}, abs/1907.11692.

\bibitem[{Misra et~al.(2020)Misra, Ettinger, and Rayz}]{misra-etal-2020-exploring}
Kanishka Misra, Allyson Ettinger, and Julia Rayz. 2020.
\newblock \href {https://doi.org/10.18653/v1/2020.findings-emnlp.415} {Exploring {BERT}{'}s sensitivity to lexical cues using tests from semantic priming}.
\newblock In \emph{Findings of the Association for Computational Linguistics: EMNLP 2020}, pages 4625--4635, Online. Association for Computational Linguistics.

\bibitem[{Nilsson(2006)}]{Nilsson2006OntologicalCF}
J{\o}rgen~Fischer Nilsson. 2006.
\newblock \href {https://doi.org/10.1007/11787181_4} {Ontological constitutions for classes and properties}.
\newblock In \emph{International Conference on Conceptual Structures}.

\bibitem[{Peng et~al.(2022)Peng, Wang, Hu, Jin, Hou, Li, Liu, and Liu}]{peng2022copen}
Hao Peng, Xiaozhi Wang, Shengding Hu, Hailong Jin, Lei Hou, Juanzi Li, Zhiyuan Liu, and Qun Liu. 2022.
\newblock \href {http://arxiv.org/abs/2211.04079} {Copen: Probing conceptual knowledge in pre-trained language models}.
\newblock In \emph{Proceedings of EMNLP}.

\bibitem[{Petroni et~al.(2020)Petroni, Lewis, Piktus, Rockt{\"a}schel, Wu, Miller, and Riedel}]{petroni-2020-context}
Fabio Petroni, Patrick Lewis, Aleksandra Piktus, Tim Rockt{\"a}schel, Yuxiang Wu, Alexander~H Miller, and Sebastian Riedel. 2020.
\newblock \href {https://doi.org/10.48550/ARXIV.2005.04611} {How context affects language models' factual predictions}.
\newblock \emph{arXiv preprint arXiv:2005.04611}.

\bibitem[{Petroni et~al.(2019)Petroni, Rockt{\"a}schel, Riedel, Lewis, Bakhtin, Wu, and Miller}]{petroni-etal-2019-language}
Fabio Petroni, Tim Rockt{\"a}schel, Sebastian Riedel, Patrick Lewis, Anton Bakhtin, Yuxiang Wu, and Alexander Miller. 2019.
\newblock \href {https://doi.org/10.18653/v1/D19-1250} {Language models as knowledge bases?}
\newblock In \emph{Proceedings of the 2019 Conference on Empirical Methods in Natural Language Processing and the 9th International Joint Conference on Natural Language Processing (EMNLP-IJCNLP)}, pages 2463--2473, Hong Kong, China. Association for Computational Linguistics.

\bibitem[{Pimentel et~al.(2020)Pimentel, Valvoda, Maudslay, Zmigrod, Williams, and Cotterell}]{pimentel-etal-2020-information}
Tiago Pimentel, Josef Valvoda, Rowan~Hall Maudslay, Ran Zmigrod, Adina Williams, and Ryan Cotterell. 2020.
\newblock \href {https://doi.org/10.18653/v1/2020.acl-main.420} {Information-theoretic probing for linguistic structure}.
\newblock In \emph{Proceedings of the 58th Annual Meeting of the Association for Computational Linguistics}, pages 4609--4622, Online. Association for Computational Linguistics.

\bibitem[{Safavi and Koutra(2021)}]{safavi-koutra-2021-relational}
Tara Safavi and Danai Koutra. 2021.
\newblock \href {https://doi.org/10.18653/v1/2021.emnlp-main.81} {{R}elational {W}orld {K}nowledge {R}epresentation in {C}ontextual {L}anguage {M}odels: {A} {R}eview}.
\newblock In \emph{Proceedings of the 2021 Conference on Empirical Methods in Natural Language Processing}, pages 1053--1067, Online and Punta Cana, Dominican Republic. Association for Computational Linguistics.

\bibitem[{Schick and Sch{\"u}tze(2021)}]{schick-schutze-2021-exploiting}
Timo Schick and Hinrich Sch{\"u}tze. 2021.
\newblock \href {https://doi.org/10.18653/v1/2021.eacl-main.20} {Exploiting cloze-questions for few-shot text classification and natural language inference}.
\newblock In \emph{Proceedings of the 16th Conference of the European Chapter of the Association for Computational Linguistics: Main Volume}, pages 255--269, Online. Association for Computational Linguistics.

\bibitem[{Sch{\"u}tze(1998)}]{schutze-1998-automatic}
Hinrich Sch{\"u}tze. 1998.
\newblock \href {https://aclanthology.org/J98-1004} {Automatic word sense discrimination}.
\newblock \emph{Computational Linguistics}, 24(1):97--123.

\bibitem[{Talmor et~al.(2019)Talmor, Herzig, Lourie, and Berant}]{talmor-etal-2019-commonsenseqa}
Alon Talmor, Jonathan Herzig, Nicholas Lourie, and Jonathan Berant. 2019.
\newblock \href {https://doi.org/10.18653/v1/N19-1421} {{C}ommonsense{QA}: A question answering challenge targeting commonsense knowledge}.
\newblock In \emph{Proceedings of the 2019 Conference of the North {A}merican Chapter of the Association for Computational Linguistics: Human Language Technologies, Volume 1 (Long and Short Papers)}, pages 4149--4158, Minneapolis, Minnesota. Association for Computational Linguistics.

\bibitem[{Talmor et~al.(2020)Talmor, Tafjord, Clark, Goldberg, and Berant}]{NEURIPS2020_e992111e}
Alon Talmor, Oyvind Tafjord, Peter Clark, Yoav Goldberg, and Jonathan Berant. 2020.
\newblock \href {https://proceedings.neurips.cc/paper/2020/file/e992111e4ab9985366e806733383bd8c-Paper.pdf} {Leap-of-thought: Teaching pre-trained models to systematically reason over implicit knowledge}.
\newblock In \emph{Advances in Neural Information Processing Systems}, volume~33, pages 20227--20237. Curran Associates, Inc.

\bibitem[{Vrandeči{\'c} and Kr{\"o}tzsch(2014)}]{Vrandei2014WikidataAF}
Denny Vrandeči{\'c} and Markus Kr{\"o}tzsch. 2014.
\newblock \href {https://doi.org/10.1145/2629489} {Wikidata: a free collaborative knowledgebase}.
\newblock \emph{Commun. ACM}, 57(10):78–85.

\bibitem[{Vuli{\'c} et~al.(2020)Vuli{\'c}, Ponti, Litschko, Glava{\v{s}}, and Korhonen}]{vulic-etal-2020-probing}
Ivan Vuli{\'c}, Edoardo~Maria Ponti, Robert Litschko, Goran Glava{\v{s}}, and Anna Korhonen. 2020.
\newblock \href {https://doi.org/10.18653/v1/2020.emnlp-main.586} {Probing pretrained language models for lexical semantics}.
\newblock In \emph{Proceedings of the 2020 Conference on Empirical Methods in Natural Language Processing (EMNLP)}, pages 7222--7240, Online. Association for Computational Linguistics.

\bibitem[{Wang et~al.(2017)Wang, Mao, Wang, and Guo}]{8047276}
Quan Wang, Zhendong Mao, Bin Wang, and Li~Guo. 2017.
\newblock \href {https://doi.org/10.1109/TKDE.2017.2754499} {Knowledge graph embedding: A survey of approaches and applications}.
\newblock \emph{IEEE Transactions on Knowledge and Data Engineering}, 29(12):2724--2743.

\bibitem[{Wang et~al.(2022)Wang, Wei, Schuurmans, Le, Chi, and Zhou}]{Wang2022SelfConsistencyIC}
Xuezhi Wang, Jason Wei, Dale Schuurmans, Quoc Le, Ed~Chi, and Denny Zhou. 2022.
\newblock \href {https://doi.org/10.48550/ARXIV.2203.11171} {Self-consistency improves chain of thought reasoning in language models}.
\newblock \emph{ArXiv}, abs/2203.11171.

\bibitem[{Wei et~al.(2022)Wei, Wang, Schuurmans, Bosma, Chi, Le, and Zhou}]{Wei2022ChainOT}
Jason Wei, Xuezhi Wang, Dale Schuurmans, Maarten Bosma, Ed~Chi, Quoc Le, and Denny Zhou. 2022.
\newblock \href {https://doi.org/10.48550/ARXIV.2201.11903} {Chain of thought prompting elicits reasoning in large language models}.
\newblock \emph{ArXiv}, abs/2201.11903.

\bibitem[{Zhang and Pei(2022)}]{Zhang2022WordKD}
Haomin~(Stanley) Zhang and Zhenxia Pei. 2022.
\newblock \href {https://link.springer.com/content/pdf/10.1007/s10936-021-09831-x} {Word knowledge dimensions in l2 lexical inference: Testing vocabulary knowledge and partial word knowledge}.
\newblock \emph{Journal of Psycholinguistic Research}, 51:151--168.

\end{thebibliography}
\bibliographystyle{acl_natbib}

\appendix
\section{Experimental Setup}
\label{appendix:setup}
We train soft tokens for 100 epochs with AdamW optimizer. The learning rate is set to 0.5 and a linear warmup scheduler is used. Since both the memorizing and reasoning task can be formulated as a multi-label classification problem, we use BCEWithLogitsLoss or NLLLoss as our loss function in the memorizing task to report the better results given by one of these two and select a better training objective. Therefore, we fix the loss function to BCEWithLogitsLoss in the reasoning task. 

For pseudowords, we set the coefficient $\alpha$ to 0.5 and sample 10 pairs of pseudowords for each entailment rule as we at most need two pseudowords to substitute the subject and object instances respectively, and report the averaged performance as the final result.

\section{Multi-token Prompting Methods}
\label{appendix:prompt}
In the main body of the paper, we discuss the impact of different \textbf{prompts} on the performance of knowledge probing and reasoning. In this section, we continuously discuss the impact of other prompt settings by comparing the averaged performance.

\subsection{Number of [MASK] Tokens} 

\begin{figure}[htbp]
    \centering
    \includegraphics[width=0.5\textwidth]{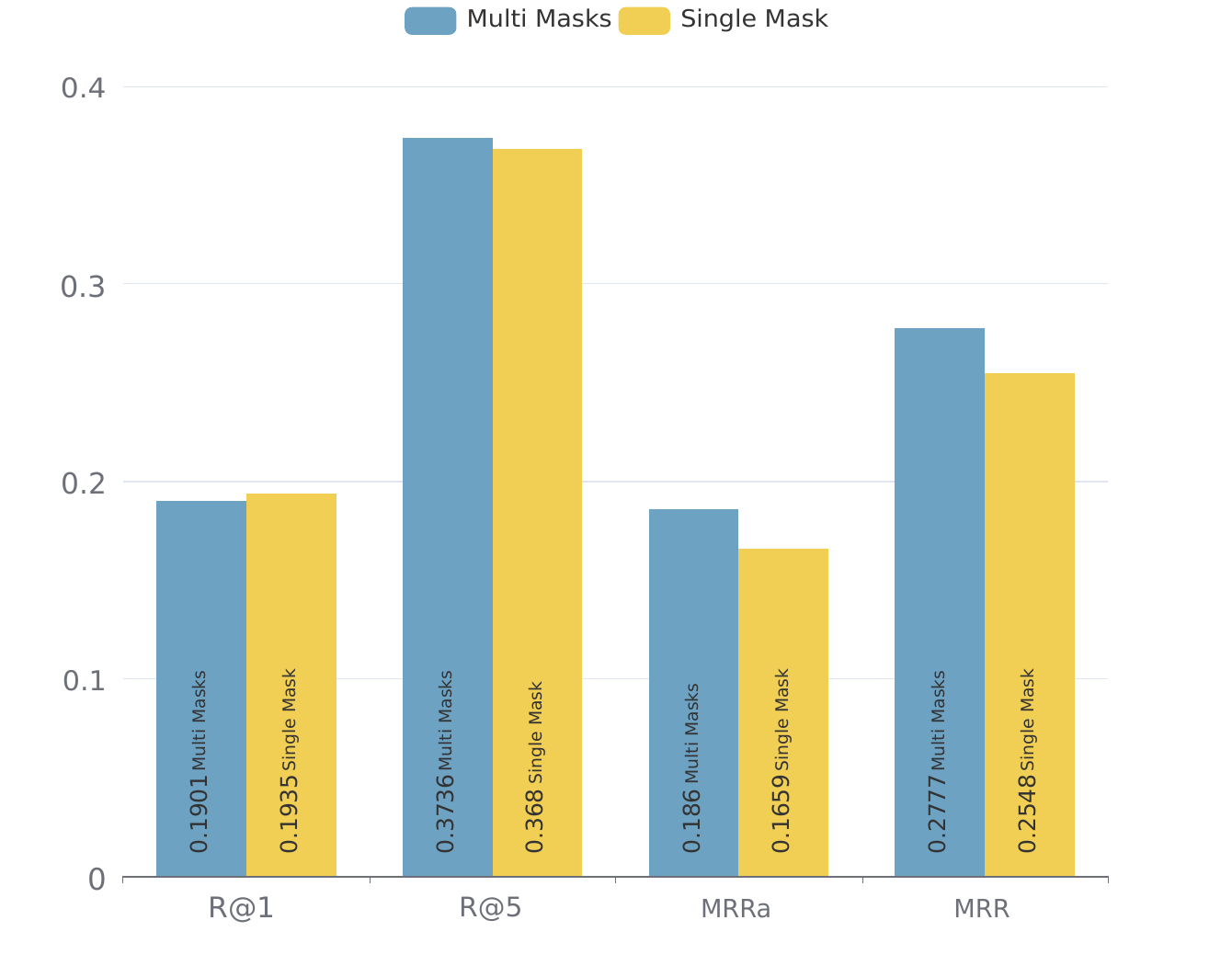}
    \caption{Comparison between multiple [MASK] tokens and a single [MASK] token in the memorizing task.}
    \label{mask}
\end{figure}

To support multi-token candidate scoring, we use multiple [MASK] tokens or one single [MASK] token to predict with masked language models. The comparison between the two methods is shown in Figure~\ref{mask}, by averaging the performance of all the memorizing tasks and models. We can observe that single [MASK] prediction achieves better accuracy (R@1) with a negligible tiny margin but worse performance in other metrics. Therefore, using multiple [MASK] tokens to obtain prediction by forward pass inference is more sensible and achieves better results.

\subsection{Pooling Methods} Three pooling methods are proposed when computing the probability of a candidate that can be tokenized into multiple subtokens. The mean-pooling method is usually used in multi-token probing. Furthermore, we introduce max-pooling and first-pooling, which retain the score of only one important token. They can exclude the influence of prepositions, e.g., by attending to \textit{mean} or \textit{transportation} when scoring the candidate \textit{mean of transportation}, but at the cost of other useful information. We are interested in whether it is better to consider the whole word or focus on the important part. 

\begin{figure}[htbp]
    \centering
    \includegraphics[width=0.5\textwidth]{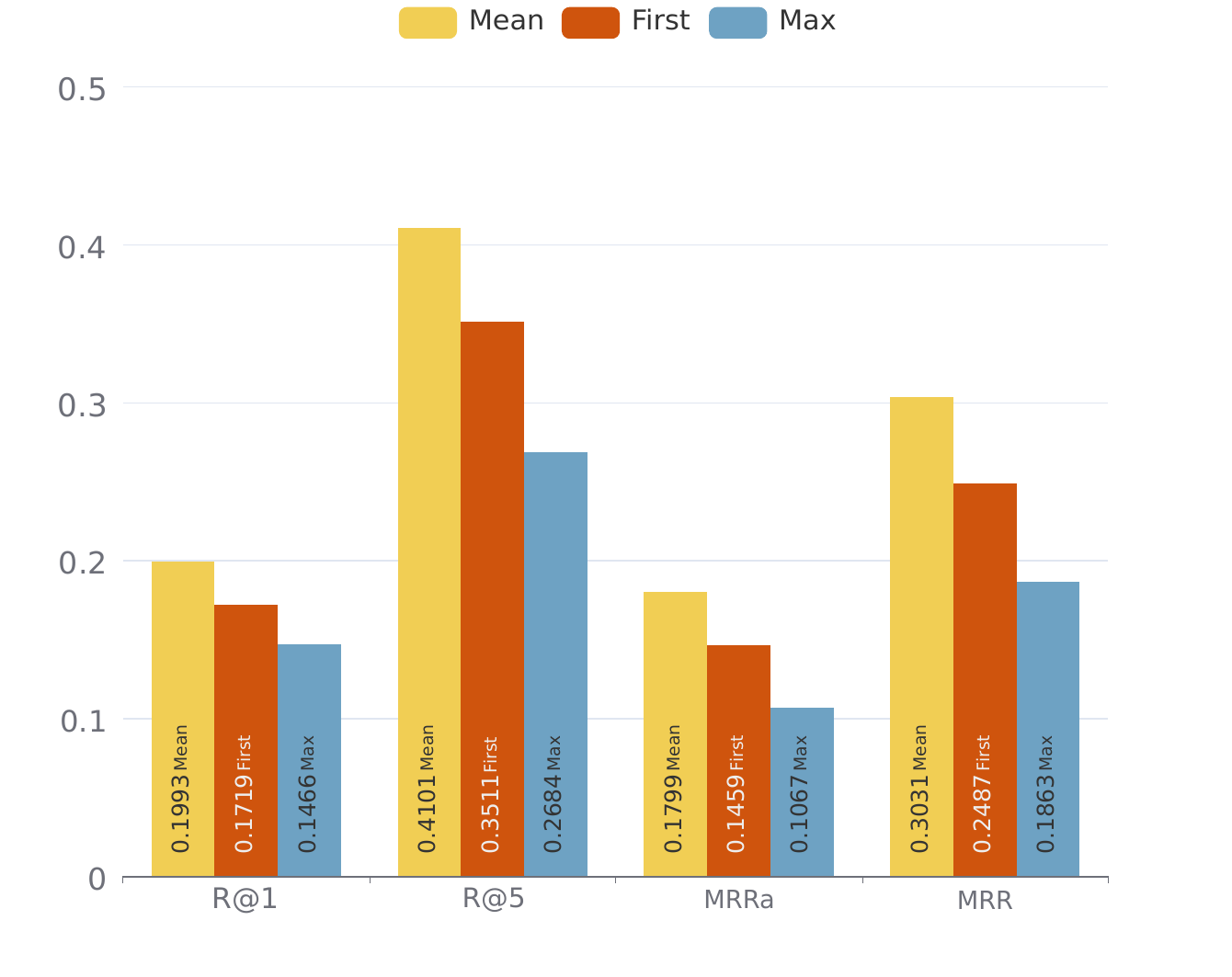}
    \caption{Effectiveness of different pooling methods in the memorizing task.}
    \label{pooling}
\end{figure}

Figure~\ref{pooling} shows that mean-pooling, as a classical method, is much better than the other two pooling methods. Besides, first-pooling gives clearly better results than max-pooling, which is possibly caused by the unique information contained in the headword (usually the first token). Consider candidates \textit{volleyball player}, \textit{squash player} and \textit{golf player}, the conditional log probability of token \textit{player} might be higher, but the candidates are distinguished by their headwords. In summary, mean-pooling obtains the best results with the most comprehensive information.

\subsection{Loss Functions} 
\begin{figure}[htbp]
    \centering
    \includegraphics[width=0.5\textwidth]{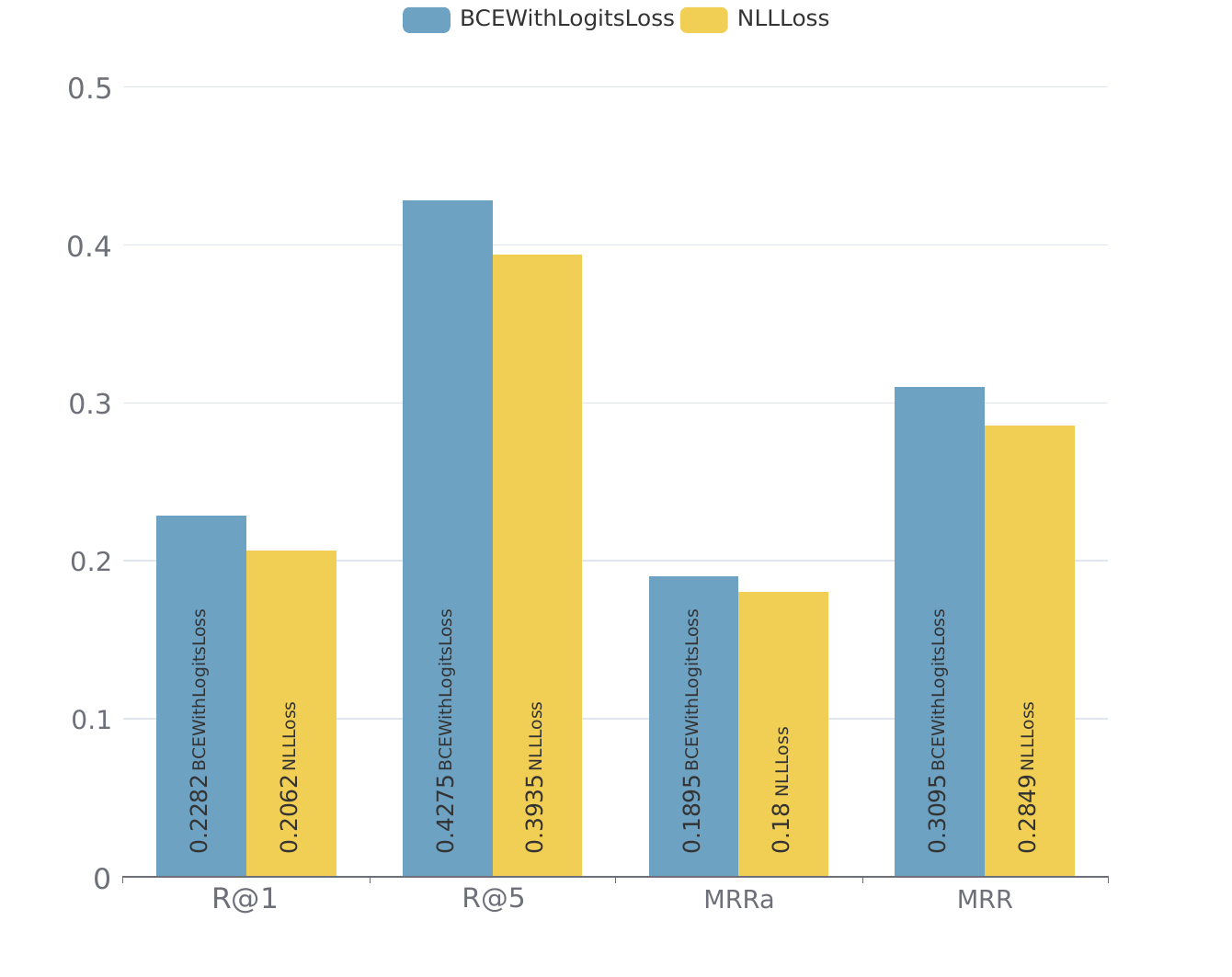}
    \caption{Comparison between two different training objectives in the memorizing task.}
    \label{loss}
\end{figure}

 As mentioned in Appendix~\ref{appendix:setup}, we try two loss functions in the memorizing task. (1) The Binary Cross Entropy With Logits Loss (BCEWithLogitsLoss) is a common loss function for multi-label classification which numerically stably combines a Sigmoid layer and the Binary Cross Entropy Loss into one layer. All examples are given the same weight when calculating the loss. (2) The Negative Log Likelihood Loss (NLLLoss) is a loss function for multi-class classification. However, we can convert the original multi-label problem to a multi-class one by sampling one ground truth at a time to generate multiple single-label multi-class classification cases. As can be seen from Figure~\ref{loss}, using BCEWithLogitsLoss as the loss function achieves better results than using NLLLoss. Hence, in subsequent reasoning experiments, we stick to the classical loss for multi-label classification. 

\section{Experimental Results}
\label{sec:appendix}

\subsection{Task Examples}

\begin{table*}[ht]
\centering
\renewcommand{\arraystretch}{0.6}
\scalebox{0.8}{
\begin{tabular}{cccc}
\toprule[1.5pt]
\multicolumn{4}{c}{Memorizing Task}                                      \\ \midrule
\multicolumn{1}{c|}{Task} & \multicolumn{1}{c|}{Prompt} & \multicolumn{1}{c|}{Top-5 Predictions} & Golds \\ \hline

\multicolumn{1}{c|}{TP} & \multicolumn{1}{c|}{Salininema is a particular [MASK] .} & \multicolumn{1}{c|}{\makecell[l]{\XSolidBrush disease\\ \XSolidBrush medical specialty\\ \XSolidBrush case\\ \XSolidBrush drug\\ \Checkmark\textcolor{Green}{species}}} & \makecell[l]{bacteria\\\textcolor{Green}{species}} \\ \hline

\multicolumn{1}{c|}{SCO} & \multicolumn{1}{c|}{Motor race is a particular [MASK] .} & \multicolumn{1}{c|}{\makecell[l]{\XSolidBrush sport\\\Checkmark \textcolor{Green}{sports event}\\\XSolidBrush genre\\\Checkmark \textcolor{Green}{event}\\\XSolidBrush  team sport}} & \makecell[l]{tournament\\\textcolor{Green}{sports event}\\societal event\\\textcolor{Green}{event}} \\ \hline

\multicolumn{1}{c|}{SPO} & \multicolumn{1}{c|}{Chief executive officer implies [MASK] .} & \multicolumn{1}{c|}{\makecell[l]{\Checkmark \textcolor{Green}{corporate officer}\\\Checkmark \textcolor{Green}{director / manager}\\\XSolidBrush significant person\\\XSolidBrush head of government\\\XSolidBrush rector}} & \makecell[l]{\textcolor{Green}{corporate officer}\\\textcolor{Green}{director / manager}}\\ \hline

\multicolumn{1}{c|}{DM} & \multicolumn{1}{c|}{One has to be a particular [MASK] to have composer.} & \multicolumn{1}{c|}{\makecell[l]{\XSolidBrush music composer\\\XSolidBrush person\\\XSolidBrush musical artist\\\XSolidBrush place\\\XSolidBrush case}} & \multicolumn{1}{l}{work} \\ \hline

\multicolumn{1}{c|}{RG} & \multicolumn{1}{c|}{One has to be a particular [MASK] to be mother.} & \multicolumn{1}{c|}{\makecell[l]{\XSolidBrush person\\\Checkmark \textcolor{Green}{woman}\\\XSolidBrush family\\\XSolidBrush name\\\XSolidBrush case}} & \multicolumn{1}{l}{\textcolor{Green}{woman}} \\ \bottomrule[1.5pt]
\end{tabular}}
\caption{Example manual prompt and predictions by BERT-base-cased for each memorizing task. Correct predictions and golds predicted among the top-5 are marked with a \Checkmark and highlighted in \textcolor{Green}{green}.}
\label{mem-ex}
\end{table*}

In order to enhance the clarity of the experiments, we have compiled a list in Table~\ref{mem-ex} that includes task prompts as well as the top five predicted candidate words generated by BERT-base-cased. The table consists of examples with successful predictions for all correct answers (SPO, RG), examples with partial correct answers predicted (TP, SCO), and examples where the correct answer is not predicted within the top five candidates (DM).

\subsection{Memorizing Results}
The complete results of the memorizing task are reported in Table~\ref{TP},~\ref{SCO},~\ref{SPO},~\ref{DM} and~\ref{RG}. 

\begin{table*}[htbp]
\scalebox{0.7}{%
\begin{tabular}{cccc|cccc|cccc|cccc}
\toprule[1.5pt]
 &  &  &  & \multicolumn{4}{c}{BERT-BASE-CASED} & \multicolumn{4}{c}{BERT-BASE-UNCASED} & \multicolumn{4}{c}{RoBERTa-BASE} \\ \midrule
\multicolumn{1}{c|}{Template} & \multicolumn{1}{c|}{Masks} & \multicolumn{1}{c|}{Pooling} & Loss & R@1 & R@5 & MRRa & MRR & R@1 & R@5 & MRRa & MRR & R@1 & R@5 & MRRa & MRR \\ \midrule
\multicolumn{1}{c|}{soft} & \multicolumn{1}{c|}{\multirow{15}{*}{m}} & \multicolumn{1}{c|}{\multirow{5}{*}{first}} & log & 18.17 & 45.45 & 1.73 & 31.18 & 18.58 & 42.26 & 1.67 & 28.83 & 7.67 & 17.00 & 0.75 & 13.46 \\
\multicolumn{1}{c|}{soft} & \multicolumn{1}{c|}{} & \multicolumn{1}{c|}{} & NLL & 20.14 & 43.13 & 1.79 & 29.94 & 19.15 & 37.91 & 1.71 & 27.08 & 8.78 & 21.19 & 0.74 & 15.95 \\
\multicolumn{1}{c|}{manual1} & \multicolumn{1}{c|}{} & \multicolumn{1}{c|}{} & \multirow{3}{*}{-} & 1.37 & 4.22 & 0.55 & 4.74 & 2.24 & 9.66 & 0.57 & 6.95 & 1.89 & 8.78 & 0.35 & 6.85 \\
\multicolumn{1}{c|}{manual2} & \multicolumn{1}{c|}{} & \multicolumn{1}{c|}{} &  & 14.22 & 31.06 & 1.07 & 22.86 & 17.23 & 34.47 & 1.17 & 25.74 & 5.03 & 11.57 & 0.49 & 9.88 \\
\multicolumn{1}{c|}{manual3} & \multicolumn{1}{c|}{} & \multicolumn{1}{c|}{} &  & 13.86 & 34.43 & 1.24 & 24.15 & 18.03 & 38.23 & 2.01 & 28.81 & 0.32 & 14.43 & 0.51 & 8.33 \\ \cmidrule{1-1} \cmidrule{3-16} 
\multicolumn{1}{c|}{soft} & \multicolumn{1}{c|}{} & \multicolumn{1}{c|}{\multirow{5}{*}{max}} & log & 15.36 & 30.73 & 1.78 & 23.26 & 12.24 & 28.19 & 1.54 & 20.24 & 10.49 & 24.51 & 1.07 & 18.31 \\
\multicolumn{1}{c|}{soft} & \multicolumn{1}{c|}{} & \multicolumn{1}{c|}{} & NLL & 10.49 & 23.67 & 1.47 & 16.79 & 15.21 & 30.42 & 1.52 & 22.08 & 1.14 & 4.61 & 0.39 & 3.35 \\
\multicolumn{1}{c|}{manual1} & \multicolumn{1}{c|}{} & \multicolumn{1}{c|}{} & \multirow{3}{*}{-} & 1.12 & 4.35 & 0.59 & 3.15 & 0.88 & 2.58 & 0.59 & 3.23 & 1.35 & 5.91 & 0.36 & 3.88 \\
\multicolumn{1}{c|}{manual2} & \multicolumn{1}{c|}{} & \multicolumn{1}{c|}{} &  & 14.06 & 26.45 & 1.15 & 18.95 & 17.31 & 32.65 & 1.23 & 23.43 & 2.28 & 7.16 & 0.40 & 4.81 \\
\multicolumn{1}{c|}{manual3} & \multicolumn{1}{c|}{} & \multicolumn{1}{c|}{} &  & 4.16 & 9.93 & 0.88 & 7.43 & 12.79 & 24.41 & 1.73 & 17.69 & 1.51 & 7.02 & 0.40 & 3.77 \\ \cmidrule{1-1} \cmidrule{3-16} 
\multicolumn{1}{c|}{soft} & \multicolumn{1}{c|}{} & \multicolumn{1}{c|}{\multirow{5}{*}{mean}} & log & 16.48 & 44.74 & 1.72 & 29.48 & 24.80 & 45.35 & 2.28 & 35.07 & 15.94 & 41.07 & 1.88 & 28.11 \\
\multicolumn{1}{c|}{soft} & \multicolumn{1}{c|}{} & \multicolumn{1}{c|}{} & NLL & 14.32 & 46.38 & 1.62 & 28.74 & 17.70 & 45.55 & 2.26 & 30.53 & 3.50 & 9.93 & 0.64 & 8.50 \\
\multicolumn{1}{c|}{manual1} & \multicolumn{1}{c|}{} & \multicolumn{1}{c|}{} & \multirow{3}{*}{-} & 9.48 & 23.19 & 1.21 & 17.05 & 4.14 & 14.81 & 0.86 & 10.18 & 2.42 & 11.67 & 0.47 & 8.65 \\
\multicolumn{1}{c|}{manual2} & \multicolumn{1}{c|}{} & \multicolumn{1}{c|}{} &  & 18.94 & 36.73 & 1.74 & 28.19 & 21.20 & 40.07 & 1.67 & 30.45 & 3.91 & 12.07 & 0.83 & 9.85 \\
\multicolumn{1}{c|}{manual3} & \multicolumn{1}{c|}{} & \multicolumn{1}{c|}{} &  & 16.21 & 41.04 & 2.01 & 28.42 & 20.84 & 45.59 & 3.14 & 33.19 & 3.63 & 8.53 & 0.84 & 8.06 \\ \midrule
\multicolumn{1}{c|}{soft} & \multicolumn{1}{c|}{\multirow{15}{*}{s}} & \multicolumn{1}{c|}{\multirow{5}{*}{first}} & log & 18.68 & 46.05 & 1.69 & 29.57 & 7.01 & 18.07 & 0.82 & 13.41 & 8.72 & 20.26 & 1.04 & 15.59 \\
\multicolumn{1}{c|}{soft} & \multicolumn{1}{c|}{} & \multicolumn{1}{c|}{} & NLL & 9.14 & 25.36 & 1.29 & 17.27 & 7.17 & 18.41 & 0.82 & 13.46 & 8.29 & 18.61 & 0.83 & 14.18 \\
\multicolumn{1}{c|}{manual1} & \multicolumn{1}{c|}{} & \multicolumn{1}{c|}{} & \multirow{3}{*}{-} & 1.73 & 5.64 & 0.62 & 6.19 & 1.24 & 9.86 & 0.65 & 6.66 & 0.43 & 4.05 & 0.37 & 4.04 \\
\multicolumn{1}{c|}{manual2} & \multicolumn{1}{c|}{} & \multicolumn{1}{c|}{} &  & 15.69 & 29.00 & 1.17 & 23.00 & 17.02 & 31.48 & 1.04 & 24.11 & 2.15 & 8.84 & 0.47 & 7.39 \\
\multicolumn{1}{c|}{manual3} & \multicolumn{1}{c|}{} & \multicolumn{1}{c|}{} &  & 12.65 & 34.11 & 1.14 & 24.10 & 17.26 & 36.81 & 1.44 & 26.62 & 2.37 & 18.25 & 0.51 & 10.83 \\ \cmidrule{1-1} \cmidrule{3-16} 
\multicolumn{1}{c|}{soft} & \multicolumn{1}{c|}{} & \multicolumn{1}{c|}{\multirow{5}{*}{max}} & log & 9.69 & 27.90 & 1.61 & 17.13 & 13.88 & 26.89 & 1.89 & 19.63 & 8.86 & 25.13 & 1.10 & 18.09 \\
\multicolumn{1}{c|}{soft} & \multicolumn{1}{c|}{} & \multicolumn{1}{c|}{} & NLL & 15.44 & 30.74 & 1.87 & 19.62 & 13.61 & 24.45 & 1.69 & 18.13 & 4.19 & 15.26 & 0.88 & 11.39 \\
\multicolumn{1}{c|}{manual1} & \multicolumn{1}{c|}{} & \multicolumn{1}{c|}{} & \multirow{3}{*}{-} & 1.12 & 3.74 & 0.79 & 4.39 & 0.94 & 3.86 & 0.76 & 4.05 & 0.74 & 5.38 & 0.37 & 2.83 \\
\multicolumn{1}{c|}{manual2} & \multicolumn{1}{c|}{} & \multicolumn{1}{c|}{} &  & 17.51 & 29.89 & 1.68 & 22.28 & 19.54 & 33.29 & 1.60 & 24.05 & 3.19 & 9.83 & 0.56 & 7.08 \\
\multicolumn{1}{c|}{manual3} & \multicolumn{1}{c|}{} & \multicolumn{1}{c|}{} &  & 11.87 & 23.55 & 1.34 & 17.52 & 15.41 & 24.45 & 1.99 & 18.25 & 1.21 & 10.75 & 0.42 & 5.59 \\ \cmidrule{1-1} \cmidrule{3-16} 
\multicolumn{1}{c|}{soft} & \multicolumn{1}{c|}{} & \multicolumn{1}{c|}{\multirow{5}{*}{mean}} & log & 10.32 & 28.26 & 1.29 & 19.91 & 13.96 & 42.95 & 2.47 & 27.58 & 4.08 & 29.67 & 1.03 & 16.81 \\
\multicolumn{1}{c|}{soft} & \multicolumn{1}{c|}{} & \multicolumn{1}{c|}{} & NLL & 10.32 & 28.29 & 1.28 & 19.93 & 21.74 & 49.28 & 2.73 & 34.51 & 3.02 & 21.65 & 0.98 & 13.03 \\
\multicolumn{1}{c|}{manual1} & \multicolumn{1}{c|}{} & \multicolumn{1}{c|}{} & \multirow{3}{*}{-} & 9.89 & 24.03 & 1.16 & 17.60 & 5.04 & 19.41 & 1.16 & 12.89 & 2.42 & 6.20 & 0.58 & 5.59 \\
\multicolumn{1}{c|}{manual2} & \multicolumn{1}{c|}{} & \multicolumn{1}{c|}{} &  & 17.02 & 32.54 & 1.48 & 25.59 & 18.72 & 35.53 & 1.38 & 27.41 & 3.29 & 12.08 & 0.88 & 8.84 \\
\multicolumn{1}{c|}{manual3} & \multicolumn{1}{c|}{} & \multicolumn{1}{c|}{} &  & 14.29 & 39.00 & 1.51 & 27.17 & 19.31 & 48.80 & 2.27 & 33.11 & 6.62 & 14.97 & 0.92 & 11.93 \\ \midrule
\multicolumn{1}{c|}{} & \multicolumn{1}{c|}{} & \multicolumn{1}{c|}{} &  & \multicolumn{4}{c}{BERT-LARGE-CASED} & \multicolumn{4}{c}{BERT-LARGE-UNCASED} & \multicolumn{4}{c}{RoBERTa-LARGE} \\ \midrule
\multicolumn{1}{c|}{soft} & \multicolumn{1}{c|}{\multirow{15}{*}{m}} & \multicolumn{1}{c|}{\multirow{5}{*}{first}} & log & 20.98 & 44.71 & 1.77 & 31.90 & 13.02 & 35.45 & 1.21 & 23.77 & 6.62 & 14.97 & 0.92 & 11.93 \\
\multicolumn{1}{c|}{soft} & \multicolumn{1}{c|}{} & \multicolumn{1}{c|}{} & NLL & 13.82 & 37.63 & 1.36 & 24.30 & 6.74 & 19.30 & 0.80 & 13.81 & 6.95 & 16.62 & 0.80 & 12.72 \\
\multicolumn{1}{c|}{manual1} & \multicolumn{1}{c|}{} & \multicolumn{1}{c|}{} & \multirow{3}{*}{-} & 2.97 & 10.44 & 0.65 & 8.52 & 2.40 & 9.74 & 0.64 & 7.66 & 5.10 & 12.25 & 0.62 & 8.53 \\
\multicolumn{1}{c|}{manual2} & \multicolumn{1}{c|}{} & \multicolumn{1}{c|}{} &  & 12.55 & 28.38 & 1.07 & 20.93 & 16.99 & 34.93 & 0.99 & 25.56 & 0.94 & 4.07 & 0.38 & 4.05 \\
\multicolumn{1}{c|}{manual3} & \multicolumn{1}{c|}{} & \multicolumn{1}{c|}{} &  & 5.60 & 28.60 & 1.47 & 17.64 & 5.95 & 21.95 & 1.04 & 14.27 & 6.26 & 17.18 & 0.75 & 12.86 \\ \cmidrule{1-1} \cmidrule{3-16} 
\multicolumn{1}{c|}{soft} & \multicolumn{1}{c|}{} & \multicolumn{1}{c|}{\multirow{5}{*}{max}} & log & 12.41 & 29.04 & 1.52 & 21.24 & 12.95 & 25.67 & 1.21 & 19.61 & 4.28 & 17.50 & 0.70 & 11.90 \\
\multicolumn{1}{c|}{soft} & \multicolumn{1}{c|}{} & \multicolumn{1}{c|}{} & NLL & 8.01 & 23.29 & 1.39 & 16.33 & 10.37 & 25.02 & 1.23 & 17.76 & 4.92 & 11.26 & 0.59 & 9.31 \\
\multicolumn{1}{c|}{manual1} & \multicolumn{1}{c|}{} & \multicolumn{1}{c|}{} & \multirow{3}{*}{-} & 1.58 & 8.14 & 0.63 & 5.21 & 1.46 & 4.65 & 0.71 & 4.54 & 0.65 & 3.28 & 0.35 & 3.25 \\
\multicolumn{1}{c|}{manual2} & \multicolumn{1}{c|}{} & \multicolumn{1}{c|}{} &  & 8.57 & 15.67 & 1.10 & 13.04 & 17.44 & 33.62 & 1.19 & 23.03 & 0.73 & 3.95 & 0.34 & 3.05 \\
\multicolumn{1}{c|}{manual3} & \multicolumn{1}{c|}{} & \multicolumn{1}{c|}{} &  & 4.80 & 14.32 & 1.28 & 10.74 & 6.04 & 11.58 & 0.91 & 8.45 & 2.69 & 7.99 & 0.43 & 5.96 \\ \cmidrule{1-1} \cmidrule{3-16} 
\multicolumn{1}{c|}{soft} & \multicolumn{1}{c|}{} & \multicolumn{1}{c|}{\multirow{5}{*}{mean}} & log & 22.87 & 50.55 & 1.96 & 35.98 & 9.56 & 40.98 & 1.33 & 23.90 & 1.71 & 7.13 & 0.48 & 4.22 \\
\multicolumn{1}{c|}{soft} & \multicolumn{1}{c|}{} & \multicolumn{1}{c|}{} & NLL & 11.70 & 37.72 & 1.70 & 24.28 & 13.06 & 32.71 & 1.38 & 23.46 & 5.60 & 14.10 & 0.73 & 11.02 \\
\multicolumn{1}{c|}{manual1} & \multicolumn{1}{c|}{} & \multicolumn{1}{c|}{} & \multirow{3}{*}{-} & 7.11 & 19.54 & 1.02 & 14.26 & 5.12 & 19.84 & 1.02 & 12.93 & 6.27 & 18.66 & 0.94 & 13.44 \\
\multicolumn{1}{c|}{manual2} & \multicolumn{1}{c|}{} & \multicolumn{1}{c|}{} &  & 15.72 & 33.10 & 1.73 & 24.96 & 22.29 & 42.14 & 1.66 & 32.10 & 3.33 & 10.50 & 0.57 & 8.49 \\
\multicolumn{1}{c|}{manual3} & \multicolumn{1}{c|}{} & \multicolumn{1}{c|}{} &  & 5.07 & 40.47 & 2.37 & 21.25 & 6.12 & 28.93 & 1.67 & 17.45 & 5.67 & 17.35 & 1.17 & 12.57 \\ \midrule
\multicolumn{1}{c|}{soft} & \multicolumn{1}{c|}{\multirow{15}{*}{s}} & \multicolumn{1}{c|}{\multirow{5}{*}{first}} & log & 15.56 & 40.12 & 1.57 & 25.67 & 11.91 & 29.20 & 1.05 & 19.23 & 7.37 & 17.92 & 1.13 & 14.32 \\
\multicolumn{1}{c|}{soft} & \multicolumn{1}{c|}{} & \multicolumn{1}{c|}{} & NLL & 9.66 & 19.80 & 1.08 & 15.79 & 12.53 & 32.42 & 1.00 & 22.02 & 5.13 & 14.18 & 0.79 & 10.12 \\
\multicolumn{1}{c|}{manual1} & \multicolumn{1}{c|}{} & \multicolumn{1}{c|}{} & \multirow{3}{*}{-} & 1.15 & 3.94 & 0.65 & 5.15 & 1.64 & 10.41 & 0.76 & 7.66 & 0.73 & 7.00 & 0.38 & 5.04 \\
\multicolumn{1}{c|}{manual2} & \multicolumn{1}{c|}{} & \multicolumn{1}{c|}{} &  & 13.30 & 27.43 & 1.17 & 20.81 & 16.87 & 30.79 & 1.08 & 23.74 & 1.29 & 8.32 & 0.39 & 5.92 \\
\multicolumn{1}{c|}{manual3} & \multicolumn{1}{c|}{} & \multicolumn{1}{c|}{} &  & 4.47 & 33.97 & 1.20 & 18.88 & 4.94 & 22.90 & 1.01 & 15.08 & 2.13 & 7.57 & 0.57 & 7.47 \\ \cmidrule{1-1} \cmidrule{3-16} 
\multicolumn{1}{c|}{soft} & \multicolumn{1}{c|}{} & \multicolumn{1}{c|}{\multirow{5}{*}{max}} & log & 11.05 & 20.58 & 1.78 & 14.88 & 12.69 & 27.23 & 1.59 & 17.51 & 6.52 & 39.09 & 0.70 & 23.67 \\
\multicolumn{1}{c|}{soft} & \multicolumn{1}{c|}{} & \multicolumn{1}{c|}{} & NLL & 13.60 & 22.89 & 1.82 & 17.26 & 12.95 & 22.48 & 1.78 & 16.59 & 7.38 & 22.45 & 0.85 & 14.86 \\
\multicolumn{1}{c|}{manual1} & \multicolumn{1}{c|}{} & \multicolumn{1}{c|}{} & \multirow{3}{*}{-} & 0.86 & 2.87 & 0.65 & 3.99 & 1.57 & 4.51 & 1.03 & 5.18 & 8.72 & 18.27 & 0.94 & 13.27 \\
\multicolumn{1}{c|}{manual2} & \multicolumn{1}{c|}{} & \multicolumn{1}{c|}{} &  & 13.79 & 27.15 & 1.72 & 19.71 & 20.50 & 34.07 & 1.79 & 24.28 & 0.52 & 5.36 & 0.40 & 2.81 \\
\multicolumn{1}{c|}{manual3} & \multicolumn{1}{c|}{} & \multicolumn{1}{c|}{} &  & 3.90 & 17.76 & 1.63 & 10.77 & 4.79 & 9.30 & 1.30 & 8.46 & 4.59 & 12.53 & 0.64 & 8.44 \\ \cmidrule{1-1} \cmidrule{3-16} 
\multicolumn{1}{c|}{soft} & \multicolumn{1}{c|}{} & \multicolumn{1}{c|}{\multirow{5}{*}{mean}} & log & 10.67 & 26.65 & 1.44 & 20.05 & 13.00 & 32.46 & 1.57 & 23.29 & 4.08 & 16.18 & 0.63 & 8.91 \\
\multicolumn{1}{c|}{soft} & \multicolumn{1}{c|}{} & \multicolumn{1}{c|}{} & NLL & 11.03 & 28.80 & 1.45 & 20.93 & 9.07 & 43.85 & 2.05 & 23.79 & 5.02 & 15.17 & 0.85 & 11.54 \\
\multicolumn{1}{c|}{manual1} & \multicolumn{1}{c|}{} & \multicolumn{1}{c|}{} & \multirow{3}{*}{-} & 7.65 & 21.82 & 1.24 & 15.66 & 5.61 & 22.93 & 1.40 & 14.74 & 5.71 & 15.88 & 0.83 & 12.37 \\
\multicolumn{1}{c|}{manual2} & \multicolumn{1}{c|}{} & \multicolumn{1}{c|}{} &  & 15.05 & 30.52 & 1.56 & 23.88 & 18.44 & 36.16 & 1.44 & 27.54 & 4.41 & 10.64 & 0.70 & 8.97 \\
\multicolumn{1}{c|}{manual3} & \multicolumn{1}{c|}{} & \multicolumn{1}{c|}{} &  & 4.56 & 46.30 & 1.73 & 21.91 & 5.02 & 27.90 & 1.51 & 18.08 & 3.95 & 15.44 & 1.19 & 10.87 \\ \bottomrule[1.5pt]
\end{tabular}%
}
\caption{TP results.}
\label{TP}
\end{table*}

\begin{table*}[htbp]
\scalebox{0.7}{%
\begin{tabular}{cccc|cccc|cccc|cccc}
\toprule[1.5pt]
 &  &  &  & \multicolumn{4}{c|}{BERT-BASE-CASED} & \multicolumn{4}{c|}{BERT-BASE-UNCASED} & \multicolumn{4}{c}{RoBERTa-BASE} \\ \midrule
\multicolumn{1}{c|}{Template} & \multicolumn{1}{c|}{Masks} & \multicolumn{1}{c|}{Pooling} & Loss & R@1 & R@5 & MRRa & MRR & R@1 & R@5 & MRRa & MRR & R@1 & R@5 & MRRa & MRR \\ \midrule
\multicolumn{1}{c|}{soft} & \multicolumn{1}{c|}{\multirow{15}{*}{m}} & \multicolumn{1}{c|}{\multirow{5}{*}{first}} & log & 10.27 & 38.09 & 4.62 & 21.50 & 34.81 & 48.79 & 10.49 & 42.26 & 22.25 & 37.95 & 6.14 & 29.75 \\
\multicolumn{1}{c|}{soft} & \multicolumn{1}{c|}{} & \multicolumn{1}{c|}{} & NLL & 7.70 & 30.24 & 4.66 & 17.00 & 32.52 & 49.36 & 11.10 & 41.13 & 10.41 & 33.81 & 4.19 & 21.20 \\
\multicolumn{1}{c|}{manual1} & \multicolumn{1}{c|}{} & \multicolumn{1}{c|}{} & \multirow{3}{*}{-} & 1.14 & 5.42 & 1.21 & 4.55 & 1.43 & 10.70 & 1.57 & 6.51 & 0.71 & 3.99 & 0.75 & 3.67 \\
\multicolumn{1}{c|}{manual2} & \multicolumn{1}{c|}{} & \multicolumn{1}{c|}{} &  & 8.84 & 25.82 & 2.02 & 16.82 & 6.85 & 21.26 & 2.18 & 15.15 & 9.84 & 22.11 & 2.57 & 16.72 \\
\multicolumn{1}{c|}{manual3} & \multicolumn{1}{c|}{} & \multicolumn{1}{c|}{} &  & 9.99 & 30.39 & 4.58 & 19.21 & 14.84 & 38.80 & 5.30 & 25.99 & 0.14 & 14.12 & 1.34 & 7.80 \\ \cmidrule{1-1} \cmidrule{3-16} 
\multicolumn{1}{c|}{soft} & \multicolumn{1}{c|}{} & \multicolumn{1}{c|}{\multirow{5}{*}{max}} & log & 29.10 & 45.51 & 7.74 & 35.85 & 24.25 & 39.37 & 4.74 & 31.07 & 15.55 & 32.24 & 5.09 & 23.50 \\
\multicolumn{1}{c|}{soft} & \multicolumn{1}{c|}{} & \multicolumn{1}{c|}{} & NLL & 5.14 & 25.39 & 3.85 & 12.32 & 11.84 & 33.52 & 4.45 & 19.75 & 5.56 & 9.99 & 2.30 & 8.63 \\
\multicolumn{1}{c|}{manual1} & \multicolumn{1}{c|}{} & \multicolumn{1}{c|}{} & \multirow{3}{*}{-} & 0.43 & 2.00 & 1.25 & 2.45 & 0.43 & 1.14 & 1.25 & 2.51 & 0.43 & 3.57 & 0.85 & 2.26 \\
\multicolumn{1}{c|}{manual2} & \multicolumn{1}{c|}{} & \multicolumn{1}{c|}{} &  & 9.42 & 23.97 & 2.25 & 13.44 & 7.28 & 21.11 & 2.66 & 11.32 & 1.28 & 6.13 & 1.19 & 3.96 \\
\multicolumn{1}{c|}{manual3} & \multicolumn{1}{c|}{} & \multicolumn{1}{c|}{} &  & 4.99 & 11.84 & 2.59 & 8.32 & 5.42 & 18.54 & 3.04 & 12.03 & 1.14 & 4.99 & 1.21 & 2.95 \\ \cmidrule{1-1} \cmidrule{3-16} 
\multicolumn{1}{c|}{soft} & \multicolumn{1}{c|}{} & \multicolumn{1}{c|}{\multirow{5}{*}{mean}} & log & 29.67 & 47.93 & 11.76 & 39.16 & 34.52 & 45.22 & 9.65 & 40.16 & 16.12 & 36.95 & 7.37 & 25.84 \\
\multicolumn{1}{c|}{soft} & \multicolumn{1}{c|}{} & \multicolumn{1}{c|}{} & NLL & 12.55 & 38.37 & 8.25 & 25.17 & 34.66 & 48.07 & 9.58 & 41.44 & 17.83 & 30.53 & 6.36 & 24.80 \\
\multicolumn{1}{c|}{manual1} & \multicolumn{1}{c|}{} & \multicolumn{1}{c|}{} & \multirow{3}{*}{-} & 7.13 & 19.12 & 3.25 & 14.18 & 2.57 & 13.69 & 2.13 & 9.01 & 1.00 & 3.00 & 1.30 & 3.98 \\
\multicolumn{1}{c|}{manual2} & \multicolumn{1}{c|}{} & \multicolumn{1}{c|}{} &  & 10.98 & 33.10 & 3.02 & 22.06 & 8.56 & 31.95 & 3.04 & 19.86 & 2.28 & 7.70 & 2.21 & 7.15 \\
\multicolumn{1}{c|}{manual3} & \multicolumn{1}{c|}{} & \multicolumn{1}{c|}{} &  & 9.56 & 38.09 & 5.27 & 22.68 & 15.12 & 43.51 & 6.55 & 28.98 & 2.14 & 7.70 & 2.35 & 6.94 \\ \midrule
\multicolumn{1}{c|}{soft} & \multicolumn{1}{c|}{\multirow{15}{*}{s}} & \multicolumn{1}{c|}{\multirow{5}{*}{first}} & log & 22.40 & 40.66 & 7.84 & 29.02 & 29.53 & 43.79 & 9.88 & 36.09 & 16.12 & 37.80 & 5.52 & 27.61 \\
\multicolumn{1}{c|}{soft} & \multicolumn{1}{c|}{} & \multicolumn{1}{c|}{} & NLL & 14.12 & 33.38 & 7.47 & 22.71 & 30.96 & 41.80 & 9.26 & 35.75 & 6.42 & 33.81 & 4.15 & 21.48 \\
\multicolumn{1}{c|}{manual1} & \multicolumn{1}{c|}{} & \multicolumn{1}{c|}{} & \multirow{3}{*}{-} & 2.43 & 7.70 & 1.86 & 6.54 & 0.29 & 6.85 & 1.44 & 4.92 & 0.71 & 1.57 & 0.95 & 2.64 \\
\multicolumn{1}{c|}{manual2} & \multicolumn{1}{c|}{} & \multicolumn{1}{c|}{} &  & 8.84 & 21.83 & 2.54 & 16.34 & 7.56 & 20.54 & 2.29 & 14.89 & 4.28 & 12.98 & 1.43 & 9.88 \\
\multicolumn{1}{c|}{manual3} & \multicolumn{1}{c|}{} & \multicolumn{1}{c|}{} &  & 7.28 & 26.25 & 4.61 & 17.40 & 13.98 & 33.52 & 3.80 & 21.68 & 7.28 & 13.98 & 2.10 & 12.37 \\ \cmidrule{1-1} \cmidrule{3-16} 
\multicolumn{1}{c|}{soft} & \multicolumn{1}{c|}{} & \multicolumn{1}{c|}{\multirow{5}{*}{max}} & log & 16.12 & 28.82 & 5.46 & 20.46 & 27.25 & 41.80 & 5.67 & 28.69 & 24.54 & 34.38 & 4.38 & 26.72 \\
\multicolumn{1}{c|}{soft} & \multicolumn{1}{c|}{} & \multicolumn{1}{c|}{} & NLL & 22.40 & 32.10 & 5.80 & 20.81 & 32.24 & 44.22 & 7.38 & 29.42 & 10.13 & 23.25 & 3.09 & 16.90 \\
\multicolumn{1}{c|}{manual1} & \multicolumn{1}{c|}{} & \multicolumn{1}{c|}{} & \multirow{3}{*}{-} & 0.86 & 3.00 & 1.94 & 3.53 & 0.86 & 2.28 & 1.65 & 2.79 & 0.14 & 4.71 & 1.13 & 2.16 \\
\multicolumn{1}{c|}{manual2} & \multicolumn{1}{c|}{} & \multicolumn{1}{c|}{} &  & 9.70 & 20.40 & 3.19 & 12.90 & 9.27 & 20.54 & 3.83 & 13.09 & 3.14 & 12.98 & 1.81 & 7.30 \\
\multicolumn{1}{c|}{manual3} & \multicolumn{1}{c|}{} & \multicolumn{1}{c|}{} &  & 6.99 & 14.69 & 3.95 & 10.95 & 13.98 & 21.83 & 4.76 & 14.10 & 1.57 & 12.27 & 1.35 & 6.09 \\ \cmidrule{1-1} \cmidrule{3-16} 
\multicolumn{1}{c|}{soft} & \multicolumn{1}{c|}{} & \multicolumn{1}{c|}{\multirow{5}{*}{mean}} & log & 23.11 & 42.51 & 8.80 & 32.21 & 37.95 & 55.49 & 13.29 & 46.45 & 19.83 & 41.37 & 8.50 & 29.48 \\
\multicolumn{1}{c|}{soft} & \multicolumn{1}{c|}{} & \multicolumn{1}{c|}{} & NLL & 8.13 & 25.25 & 5.52 & 17.82 & 36.09 & 55.92 & 9.90 & 45.37 & 17.55 & 36.09 & 7.25 & 26.81 \\
\multicolumn{1}{c|}{manual1} & \multicolumn{1}{c|}{} & \multicolumn{1}{c|}{} & \multirow{3}{*}{-} & 7.56 & 18.97 & 3.44 & 14.54 & 2.71 & 15.55 & 2.44 & 9.92 & 1.14 & 4.42 & 1.85 & 4.40 \\
\multicolumn{1}{c|}{manual2} & \multicolumn{1}{c|}{} & \multicolumn{1}{c|}{} &  & 9.56 & 28.67 & 3.36 & 19.58 & 8.84 & 25.39 & 3.09 & 18.75 & 2.00 & 10.27 & 2.60 & 8.06 \\
\multicolumn{1}{c|}{manual3} & \multicolumn{1}{c|}{} & \multicolumn{1}{c|}{} &  & 8.42 & 34.52 & 4.80 & 21.78 & 15.12 & 41.80 & 5.24 & 27.74 & 6.70 & 17.12 & 4.24 & 13.44 \\ \midrule
\multicolumn{1}{c|}{} & \multicolumn{1}{c|}{} & \multicolumn{1}{c|}{} &  & \multicolumn{4}{c|}{BERT-LARGE-CASED} & \multicolumn{4}{c|}{BERT-LARGE-UNCASED} & \multicolumn{4}{c}{RoBERTa-LARGE} \\ \midrule
\multicolumn{1}{c|}{soft} & \multicolumn{1}{c|}{\multirow{15}{*}{m}} & \multicolumn{1}{c|}{\multirow{5}{*}{first}} & log & 15.41 & 41.94 & 6.05 & 26.93 & 28.82 & 44.79 & 5.28 & 36.93 & 16.69 & 26.82 & 3.69 & 21.80 \\
\multicolumn{1}{c|}{soft} & \multicolumn{1}{c|}{} & \multicolumn{1}{c|}{} & NLL & 20.40 & 43.94 & 6.80 & 32.20 & 25.68 & 43.22 & 6.14 & 34.56 & 10.56 & 21.40 & 3.42 & 16.51 \\
\multicolumn{1}{c|}{manual1} & \multicolumn{1}{c|}{} & \multicolumn{1}{c|}{} & \multirow{3}{*}{-} & 4.14 & 11.70 & 1.73 & 9.27 & 2.43 & 9.70 & 1.66 & 7.29 & 0.71 & 3.99 & 1.00 & 3.53 \\
\multicolumn{1}{c|}{manual2} & \multicolumn{1}{c|}{} & \multicolumn{1}{c|}{} &  & 5.71 & 23.97 & 2.60 & 14.90 & 6.99 & 22.97 & 2.25 & 15.16 & 5.99 & 19.97 & 2.60 & 13.88 \\
\multicolumn{1}{c|}{manual3} & \multicolumn{1}{c|}{} & \multicolumn{1}{c|}{} &  & 13.98 & 37.80 & 5.56 & 24.04 & 5.42 & 18.54 & 2.46 & 11.78 & 9.13 & 26.11 & 3.02 & 17.96 \\ \cmidrule{1-1} \cmidrule{3-16} 
\multicolumn{1}{c|}{soft} & \multicolumn{1}{c|}{} & \multicolumn{1}{c|}{\multirow{5}{*}{max}} & log & 21.68 & 36.38 & 5.03 & 28.70 & 28.82 & 40.37 & 6.04 & 34.92 & 11.41 & 24.82 & 3.16 & 17.69 \\
\multicolumn{1}{c|}{soft} & \multicolumn{1}{c|}{} & \multicolumn{1}{c|}{} & NLL & 5.71 & 20.11 & 4.21 & 14.18 & 12.55 & 22.97 & 4.69 & 18.47 & 6.99 & 13.98 & 3.80 & 11.41 \\
\multicolumn{1}{c|}{manual1} & \multicolumn{1}{c|}{} & \multicolumn{1}{c|}{} & \multirow{3}{*}{-} & 1.85 & 8.70 & 1.75 & 6.51 & 1.43 & 4.99 & 2.11 & 3.99 & 0.14 & 2.43 & 0.86 & 2.13 \\
\multicolumn{1}{c|}{manual2} & \multicolumn{1}{c|}{} & \multicolumn{1}{c|}{} &  & 5.85 & 12.98 & 2.11 & 8.79 & 11.55 & 25.53 & 2.73 & 13.79 & 0.86 & 4.99 & 1.00 & 3.36 \\
\multicolumn{1}{c|}{manual3} & \multicolumn{1}{c|}{} & \multicolumn{1}{c|}{} &  & 7.13 & 21.54 & 3.78 & 15.57 & 2.57 & 8.27 & 2.57 & 5.92 & 1.14 & 5.14 & 1.16 & 3.40 \\ \cmidrule{1-1} \cmidrule{3-16} 
\multicolumn{1}{c|}{soft} & \multicolumn{1}{c|}{} & \multicolumn{1}{c|}{\multirow{5}{*}{mean}} & log & 24.25 & 38.66 & 4.99 & 31.50 & 21.40 & 42.80 & 5.04 & 32.11 & 22.68 & 42.51 & 5.54 & 32.74 \\
\multicolumn{1}{c|}{soft} & \multicolumn{1}{c|}{} & \multicolumn{1}{c|}{} & NLL & 22.40 & 42.65 & 6.04 & 32.95 & 30.96 & 53.50 & 7.99 & 41.91 & 22.82 & 42.80 & 4.77 & 32.89 \\
\multicolumn{1}{c|}{manual1} & \multicolumn{1}{c|}{} & \multicolumn{1}{c|}{} & \multirow{3}{*}{-} & 5.14 & 21.68 & 2.50 & 13.72 & 3.57 & 17.40 & 2.56 & 10.83 & 2.00 & 7.42 & 1.81 & 5.88 \\
\multicolumn{1}{c|}{manual2} & \multicolumn{1}{c|}{} & \multicolumn{1}{c|}{} &  & 8.27 & 31.38 & 3.69 & 19.77 & 9.27 & 35.38 & 3.15 & 21.89 & 2.43 & 12.55 & 2.36 & 8.65 \\
\multicolumn{1}{c|}{manual3} & \multicolumn{1}{c|}{} & \multicolumn{1}{c|}{} &  & 10.27 & 43.79 & 6.71 & 25.80 & 3.99 & 18.83 & 3.39 & 12.79 & 6.13 & 16.83 & 3.50 & 13.53 \\ \midrule
\multicolumn{1}{c|}{soft} & \multicolumn{1}{c|}{\multirow{15}{*}{s}} & \multicolumn{1}{c|}{\multirow{5}{*}{first}} & log & 30.96 & 47.65 & 7.24 & 35.95 & 24.82 & 46.08 & 8.83 & 33.06 & 6.56 & 12.70 & 2.33 & 11.33 \\
\multicolumn{1}{c|}{soft} & \multicolumn{1}{c|}{} & \multicolumn{1}{c|}{} & NLL & 34.95 & 49.22 & 8.96 & 37.90 & 25.25 & 44.08 & 8.90 & 32.90 & 9.70 & 26.25 & 2.92 & 19.40 \\
\multicolumn{1}{c|}{manual1} & \multicolumn{1}{c|}{} & \multicolumn{1}{c|}{} & \multirow{3}{*}{-} & 1.57 & 4.56 & 1.80 & 5.56 & 1.85 & 8.42 & 1.98 & 6.92 & 1.43 & 7.28 & 1.26 & 5.85 \\
\multicolumn{1}{c|}{manual2} & \multicolumn{1}{c|}{} & \multicolumn{1}{c|}{} &  & 7.56 & 20.83 & 3.15 & 16.15 & 7.42 & 23.40 & 2.61 & 15.53 & 2.43 & 7.99 & 1.41 & 7.77 \\
\multicolumn{1}{c|}{manual3} & \multicolumn{1}{c|}{} & \multicolumn{1}{c|}{} &  & 9.70 & 31.38 & 4.94 & 20.07 & 3.71 & 16.12 & 2.47 & 11.34 & 12.27 & 39.80 & 3.21 & 24.56 \\ \cmidrule{1-1} \cmidrule{3-16} 
\multicolumn{1}{c|}{soft} & \multicolumn{1}{c|}{} & \multicolumn{1}{c|}{\multirow{5}{*}{max}} & log & 32.38 & 46.65 & 8.60 & 30.89 & 20.54 & 33.10 & 5.41 & 25.27 & 12.13 & 22.82 & 3.20 & 16.22 \\
\multicolumn{1}{c|}{soft} & \multicolumn{1}{c|}{} & \multicolumn{1}{c|}{} & NLL & 32.38 & 44.22 & 7.81 & 26.46 & 0.00 & 1.43 & 0.78 & 1.63 & 8.56 & 17.69 & 2.14 & 13.28 \\
\multicolumn{1}{c|}{manual1} & \multicolumn{1}{c|}{} & \multicolumn{1}{c|}{} & \multirow{3}{*}{-} & 0.29 & 2.28 & 1.61 & 3.19 & 2.28 & 5.28 & 2.65 & 4.60 & 0.14 & 4.85 & 0.98 & 2.25 \\
\multicolumn{1}{c|}{manual2} & \multicolumn{1}{c|}{} & \multicolumn{1}{c|}{} &  & 7.99 & 18.83 & 3.97 & 12.91 & 11.13 & 24.11 & 3.72 & 13.35 & 4.14 & 12.70 & 1.96 & 6.97 \\
\multicolumn{1}{c|}{manual3} & \multicolumn{1}{c|}{} & \multicolumn{1}{c|}{} &  & 7.85 & 16.83 & 4.81 & 11.86 & 3.71 & 5.85 & 2.64 & 5.59 & 5.14 & 20.11 & 2.20 & 9.66 \\ \cmidrule{1-1} \cmidrule{3-16} 
\multicolumn{1}{c|}{soft} & \multicolumn{1}{c|}{} & \multicolumn{1}{c|}{\multirow{5}{*}{mean}} & log & 28.96 & 50.50 & 9.16 & 39.13 & 19.26 & 44.79 & 6.02 & 30.95 & 14.27 & 36.95 & 5.14 & 25.52 \\
\multicolumn{1}{c|}{soft} & \multicolumn{1}{c|}{} & \multicolumn{1}{c|}{} & NLL & 29.53 & 54.64 & 9.70 & 41.18 & 20.11 & 31.38 & 5.34 & 26.79 & 8.27 & 22.97 & 4.71 & 16.72 \\
\multicolumn{1}{c|}{manual1} & \multicolumn{1}{c|}{} & \multicolumn{1}{c|}{} & \multirow{3}{*}{-} & 5.99 & 21.54 & 2.36 & 14.33 & 3.85 & 19.83 & 3.05 & 11.94 & 1.57 & 8.27 & 2.79 & 6.34 \\
\multicolumn{1}{c|}{manual2} & \multicolumn{1}{c|}{} & \multicolumn{1}{c|}{} &  & 9.27 & 29.81 & 3.77 & 19.62 & 8.42 & 33.24 & 3.45 & 19.95 & 1.85 & 12.41 & 2.90 & 8.24 \\
\multicolumn{1}{c|}{manual3} & \multicolumn{1}{c|}{} & \multicolumn{1}{c|}{} &  & 11.55 & 37.95 & 5.68 & 24.57 & 3.71 & 24.96 & 3.42 & 14.69 & 4.56 & 21.40 & 4.45 & 13.97 \\
\bottomrule[1.5pt]
\end{tabular}%
}
\caption{SCO results.}
\label{SCO}
\end{table*}

\begin{table*}[htbp]
\scalebox{0.7}{%
\begin{tabular}{cccc|cccc|cccc|cccc}
\toprule[1.5pt]
 &  &  &  & \multicolumn{4}{c|}{BERT-BASE-CASED} & \multicolumn{4}{c|}{BERT-BASE-UNCASED} & \multicolumn{4}{c}{RoBERTa-BASE} \\ \midrule
\multicolumn{1}{c|}{Template} & \multicolumn{1}{c|}{Masks} & \multicolumn{1}{c|}{Pooling} & Loss & R@1 & R@5 & MRRa & MRR & R@1 & R@5 & MRRa & MRR & R@1 & R@5 & MRRa & MRR \\ \midrule
\multicolumn{1}{c|}{soft} & \multicolumn{1}{c|}{} & \multicolumn{1}{c|}{} & log & 20.51 & 43.59 & 15.37 & 32 & 20.51 & 61.54 & 19.41 & 36.06 & 7.69 & 43.59 & 11.31 & 20.65 \\
\multicolumn{1}{c|}{soft} & \multicolumn{1}{c|}{} & \multicolumn{1}{c|}{} & NLL & 23.08 & 38.46 & 15.44 & 33.36 & 20.51 & 58.97 & 18.61 & 37.51 & 2.56 & 43.59 & 11.09 & 21.63 \\
\multicolumn{1}{c|}{manual} & \multicolumn{1}{c|}{} & \multicolumn{1}{c|}{\multirow{-3}{*}{first}} & - & 20.51 & 58.97 & 13.5 & 34.67 & 17.95 & 48.72 & 16.15 & 32.42 & 10.26 & 25.64 & 8.77 & 20.34 \\ \cmidrule{1-1} \cmidrule{3-16} 
\multicolumn{1}{c|}{soft} & \multicolumn{1}{c|}{} & \multicolumn{1}{c|}{} & log & 23.08 & 64.1 & 20.93 & 43.68 & 28.21 & 58.97 & 25.12 & 44.43 & 12.82 & 28.21 & 12.02 & 18.46 \\
\multicolumn{1}{c|}{soft} & \multicolumn{1}{c|}{} & \multicolumn{1}{c|}{} & NLL & 20.51 & 64.1 & 21.13 & 39.21 & 38.46 & 58.97 & 22.5 & 45.98 & 15.38 & 35.9 & 15.03 & 27.21 \\
\multicolumn{1}{c|}{manual} & \multicolumn{1}{c|}{} & \multicolumn{1}{c|}{\multirow{-3}{*}{max}} & - & 7.69 & 25.64 & 9.47 & 19.56 & 7.69 & 35.9 & 9.26 & 21.12 & 0 & 10.26 & 4.51 & 7.27 \\ \cmidrule{1-1} \cmidrule{3-16} 
\multicolumn{1}{c|}{soft} & \multicolumn{1}{c|}{} & \multicolumn{1}{c|}{} & log & 17.95 & 64.1 & 20.97 & 35.62 & 38.46 & 71.79 & 29.32 & 53.51 & 35.9 & 61.54 & 22.23 & 47.35 \\
\multicolumn{1}{c|}{soft} & \multicolumn{1}{c|}{} & \multicolumn{1}{c|}{} & NLL & 25.64 & 51.28 & 21.33 & 38.26 & 28.21 & 74.36 & 25.87 & 47.12 & 33.33 & 61.54 & 25.12 & 46.47 \\
\multicolumn{1}{c|}{manual} & \multicolumn{1}{c|}{\multirow{-9}{*}{m}} & \multicolumn{1}{c|}{\multirow{-3}{*}{mean}} & - & 23.08 & 64.1 & 15.81 & 39.17 & 17.95 & 69.23 & 19.48 & 38.11 & 10.26 & 25.64 & 7.91 & 18.72 \\ \midrule
\multicolumn{1}{c|}{soft} & \multicolumn{1}{c|}{} & \multicolumn{1}{c|}{} & log & 15.38 & 35.9 & 18.38 & 29.45 & 28.21 & 58.97 & 20.01 & 34.91 & 20.51 & 35.9 & 14.42 & 28.34 \\
\multicolumn{1}{c|}{soft} & \multicolumn{1}{c|}{} & \multicolumn{1}{c|}{} & NLL & 25.64 & 41.03 & 15.8 & 31.25 & 25.64 & 51.28 & 17.82 & 33.3 & 20.51 & 43.59 & 15.67 & 26.17 \\
\multicolumn{1}{c|}{manual} & \multicolumn{1}{c|}{} & \multicolumn{1}{c|}{\multirow{-3}{*}{first}} & - & 20.51 & 53.85 & 13.12 & 33.99 & 17.95 & 61.54 & 16.75 & 35.37 & 10.26 & 33.33 & 8.35 & 20.6 \\ \cmidrule{1-1} \cmidrule{3-16} 
\multicolumn{1}{c|}{soft} & \multicolumn{1}{c|}{} & \multicolumn{1}{c|}{} & log & 30.77 & 64.1 & 22.04 & 42.89 & 20.51 & 35.9 & 12.16 & 27.34 & 20.51 & 33.33 & 12.14 & 24.43 \\
\multicolumn{1}{c|}{soft} & \multicolumn{1}{c|}{} & \multicolumn{1}{c|}{} & NLL & 38.46 & 48.72 & 21.48 & 43.04 & 17.95 & 33.33 & 12.48 & 27.17 & 20.51 & 28.21 & 11.93 & 27.11 \\
\multicolumn{1}{c|}{manual} & \multicolumn{1}{c|}{} & \multicolumn{1}{c|}{\multirow{-3}{*}{max}} & - & 20.51 & 43.59 & 10.66 & 27.05 & 15.38 & 51.28 & 16.53 & 33.61 & 0 & 7.69 & 3.28 & 8.36 \\ \cmidrule{1-1} \cmidrule{3-16} 
\multicolumn{1}{c|}{soft} & \multicolumn{1}{c|}{} & \multicolumn{1}{c|}{} & log & 30.77 & 64.1 & 23.81 & 42.82 & 23.08 & 56.41 & 21.8 & 39.4 & 33.33 & 61.54 & 23.05 & 46.92 \\
\multicolumn{1}{c|}{soft} & \multicolumn{1}{c|}{} & \multicolumn{1}{c|}{} & NLL & 20.51 & 48.72 & 17 & 33.75 & 20.51 & 56.41 & 23.15 & 37.13 & 30.77 & 61.54 & 22.35 & 44.84 \\
\multicolumn{1}{c|}{manual} & \multicolumn{1}{c|}{\multirow{-9}{*}{s}} & \multicolumn{1}{c|}{\multirow{-3}{*}{mean}} & - & 20.51 & 53.85 & 13.3 & 34.31 & 20.51 & 61.54 & 17.85 & 38.29 & 7.69 & 20.51 & 7.21 & 16.25 \\ \midrule
 &  &  &  & \multicolumn{4}{c|}{BERT-LARGE-CASED} & \multicolumn{4}{c|}{BERT-LARGE-UNCASED} & \multicolumn{4}{c}{RoBERTa-LARGE} \\ \midrule
\multicolumn{1}{c|}{soft} & \multicolumn{1}{c|}{} & \multicolumn{1}{c|}{} & log & 30.77 & 61.54 & 20.95 & 41.45 & 15.38 & 53.85 & 18.47 & 30.22 & 17.95 & 43.59 & 16.03 & 26.64 \\
\multicolumn{1}{c|}{soft} & \multicolumn{1}{c|}{} & \multicolumn{1}{c|}{} & NLL & 38.46 & 56.41 & 16.24 & 32.73 & 30.77 & 56.41 & 20.75 & 34.34 & 15.38 & 33.33 & 13.59 & 25.34 \\
\multicolumn{1}{c|}{manual} & \multicolumn{1}{c|}{} & \multicolumn{1}{c|}{\multirow{-3}{*}{first}} & - & 10.26 & 51.28 & 15.19 & 28.84 & 15.38 & 43.59 & 14.99 & 28.95 & 7.69 & 30.77 & 7.99 & 21.87 \\ \cmidrule{1-1} \cmidrule{3-16} 
\multicolumn{1}{c|}{soft} & \multicolumn{1}{c|}{} & \multicolumn{1}{c|}{} & log & 28.21 & 58.97 & 23.35 & 42.49 & 25.64 & 46.15 & 17.68 & 37.26 & 17.95 & 46.15 & 11.04 & 27.7 \\
\multicolumn{1}{c|}{soft} & \multicolumn{1}{c|}{} & \multicolumn{1}{c|}{} & NLL & 10.26 & 43.59 & 13.35 & 26.44 & 23.08 & 56.41 & 17.96 & 39.64 & 17.95 & 51.28 & 9.54 & 27.72 \\
\multicolumn{1}{c|}{manual} & \multicolumn{1}{c|}{} & \multicolumn{1}{c|}{\multirow{-3}{*}{max}} & - & 7.69 & 41.03 & 11.76 & 23.03 & 12.82 & 35.9 & 11.38 & 26.06 & 0 & 2.56 & 3.99 & 7.27 \\ \cmidrule{1-1} \cmidrule{3-16} 
\multicolumn{1}{c|}{soft} & \multicolumn{1}{c|}{} & \multicolumn{1}{c|}{} & log & 28.21 & 76.92 & 28.31 & 49.83 & 41.03 & 64.1 & 28.83 & 52.91 & 23.08 & 51.28 & 17.42 & 34.65 \\
\multicolumn{1}{c|}{soft} & \multicolumn{1}{c|}{} & \multicolumn{1}{c|}{} & NLL & 35.9 & 64.1 & 29.8 & 48.47 & 38.46 & 64.1 & 26.7 & 49.25 & 25.64 & 35.9 & 13.9 & 33.24 \\
\multicolumn{1}{c|}{manual} & \multicolumn{1}{c|}{\multirow{-9}{*}{m}} & \multicolumn{1}{c|}{\multirow{-3}{*}{mean}} & - & 10.26 & 58.97 & 19.52 & 33.82 & 20.51 & 69.23 & 18.97 & 39.31 & 5.13 & 23.08 & 7.45 & 17.07 \\ \midrule
\multicolumn{1}{c|}{soft} & \multicolumn{1}{c|}{} & \multicolumn{1}{c|}{} & log & 30.77 & 64.1 & 17.89 & 32.48 & 20.51 & 61.54 & 16.96 & 32.09 & 15.38 & 51.28 & 16.87 & 30.19 \\
\multicolumn{1}{c|}{soft} & \multicolumn{1}{c|}{} & \multicolumn{1}{c|}{} & NLL & 30.77 & 53.85 & 13.78 & 25.24 & 5.13 & 12.82 & 3.87 & 13.48 & 17.95 & 43.59 & 14.73 & 24.36 \\
\multicolumn{1}{c|}{manual} & \multicolumn{1}{c|}{} & \multicolumn{1}{c|}{\multirow{-3}{*}{first}} & - & 15.38 & 53.85 & 15.59 & 33.3 & 23.08 & 48.72 & 14.41 & 33.99 & 10.26 & 30.77 & 10.02 & 21.01 \\ \cmidrule{1-1} \cmidrule{3-16} 
\multicolumn{1}{c|}{soft} & \multicolumn{1}{c|}{} & \multicolumn{1}{c|}{} & log & 25.64 & 43.59 & 17.92 & 33.47 & 20.51 & 46.15 & 16.76 & 31.9 & 15.38 & 43.59 & 9.44 & 25.14 \\
\multicolumn{1}{c|}{soft} & \multicolumn{1}{c|}{} & \multicolumn{1}{c|}{} & NLL & 25.64 & 23.08 & 11.77 & 29.8 & 33.33 & 56.41 & 22.49 & 44.09 & 15.38 & 38.46 & 9.82 & 26.53 \\
\multicolumn{1}{c|}{manual} & \multicolumn{1}{c|}{} & \multicolumn{1}{c|}{\multirow{-3}{*}{max}} & - & 17.95 & 53.85 & 14.78 & 29.56 & 20.51 & 51.28 & 14.68 & 33.25 & 2.56 & 10.26 & 4.28 & 11.03 \\ \cmidrule{1-1} \cmidrule{3-16} 
\multicolumn{1}{c|}{soft} & \multicolumn{1}{c|}{} & \multicolumn{1}{c|}{} & log & 33.33 & 58.97 & 24.04 & 44.57 & 17.95 & 56.41 & 18.76 & 36.52 & 33.33 & 69.23 & 22.07 & 47.02 \\
\multicolumn{1}{c|}{soft} & \multicolumn{1}{c|}{} & \multicolumn{1}{c|}{} & NLL & 23.08 & 58.97 & 20.72 & 40.35 & 23.08 & 64.1 & 21.55 & 39.82 & 41.03 & 69.23 & 29.61 & 53.77 \\
\multicolumn{1}{c|}{manual} & \multicolumn{1}{c|}{\multirow{-9}{*}{s}} & \multicolumn{1}{c|}{\multirow{-3}{*}{mean}} & - & 15.38 & 56.41 & 17.15 & 34.53 & 17.95 & 53.85 & 15.81 & 34.43 & 10.26 & 20.51 & 7.43 & 19.49 \\ \bottomrule[1.5pt]
\end{tabular}%
}
\caption{SPO results.}
\label{SPO}
\end{table*}

\begin{table*}[htbp]
\scalebox{0.7}{%
\begin{tabular}{cccc|cccc|cccc|cccc}
\toprule[1.5pt]
 &  &  &  & \multicolumn{4}{c|}{BERT-BASE-CASED} & \multicolumn{4}{c|}{BERT-BASE-UNCASED} & \multicolumn{4}{c}{RoBERTa-BASE} \\ \midrule
\multicolumn{1}{c|}{Template} & \multicolumn{1}{c|}{Masks} & \multicolumn{1}{c|}{Pooling} & Loss & R@1 & R@5 & MRRa & MRR & R@1 & R@5 & MRRa & MRR & R@1 & R@5 & MRRa & MRR \\ \midrule
\multicolumn{1}{c|}{soft} & \multicolumn{1}{c|}{} & \multicolumn{1}{c|}{} & log & 30.00 & 56.67 & 39.28 & 39.28 & 10.00 & 36.67 & 23.14 & 23.14 & 20.00 & 63.33 & 39.59 & 39.59 \\
\multicolumn{1}{c|}{soft} & \multicolumn{1}{c|}{} & \multicolumn{1}{c|}{} & NLL & 3.33 & 13.33 & 11.03 & 11.03 & 6.67 & 20.00 & 10.17 & 10.17 & 20.00 & 63.33 & 38.52 & 38.52 \\
\multicolumn{1}{c|}{manual} & \multicolumn{1}{c|}{} & \multicolumn{1}{c|}{\multirow{-3}{*}{first}} & - & 40.00 & 46.67 & 44.06 & 44.06 & 43.33 & 46.67 & 46.65 & 46.65 & 0.00 & 3.33 & 3.26 & 3.26 \\ \cmidrule{1-1} \cmidrule{3-16} 
\multicolumn{1}{c|}{soft} & \multicolumn{1}{c|}{} & \multicolumn{1}{c|}{} & log & 3.33 & 10.00 & 8.38 & 8.38 & 30.00 & 43.33 & 36.96 & 36.96 & 30.00 & 43.33 & 37.14 & 37.14 \\
\multicolumn{1}{c|}{soft} & \multicolumn{1}{c|}{} & \multicolumn{1}{c|}{} & NLL & 0.00 & 0.00 & 2.45 & 2.45 & 20.00 & 26.67 & 23.66 & 23.66 & 13.33 & 16.67 & 16.62 & 16.62 \\
\multicolumn{1}{c|}{manual} & \multicolumn{1}{c|}{} & \multicolumn{1}{c|}{\multirow{-3}{*}{max}} & - & 33.33 & 46.67 & 39.34 & 39.34 & 40.00 & 46.67 & 43.32 & 43.32 & 0.00 & 0.00 & 0.46 & 0.46 \\ \cmidrule{1-1} \cmidrule{3-16} 
\multicolumn{1}{c|}{soft} & \multicolumn{1}{c|}{} & \multicolumn{1}{c|}{} & log & 23.33 & 60.00 & 40.66 & 40.66 & 40.00 & 63.33 & 50.02 & 50.02 & 40.00 & 60.00 & 49.00 & 49.00 \\
\multicolumn{1}{c|}{soft} & \multicolumn{1}{c|}{} & \multicolumn{1}{c|}{} & NLL & 13.33 & 46.67 & 29.67 & 29.67 & 30.00 & 43.33 & 38.77 & 38.77 & 13.33 & 53.33 & 32.36 & 32.36 \\
\multicolumn{1}{c|}{manual} & \multicolumn{1}{c|}{\multirow{-9}{*}{m}} & \multicolumn{1}{c|}{\multirow{-3}{*}{mean}} & - & 43.33 & 50.00 & 46.91 & 46.91 & 43.33 & 53.33 & 48.65 & 48.65 & 0.00 & 3.33 & 4.00 & 4.00 \\ \midrule
\multicolumn{1}{c|}{soft} & \multicolumn{1}{c|}{} & \multicolumn{1}{c|}{} & log & 16.67 & 53.33 & 32.13 & 32.13 & 20.00 & 50.00 & 27.56 & 27.56 & 10.00 & 30.00 & 18.79 & 18.79 \\
\multicolumn{1}{c|}{soft} & \multicolumn{1}{c|}{} & \multicolumn{1}{c|}{} & NLL & 13.33 & 43.33 & 27.50 & 27.50 & 13.33 & 36.67 & 25.53 & 25.53 & 10.00 & 26.67 & 18.12 & 18.12 \\
\multicolumn{1}{c|}{manual} & \multicolumn{1}{c|}{} & \multicolumn{1}{c|}{\multirow{-3}{*}{first}} & - & 43.33 & 50.00 & 36.40 & 36.40 & 40.00 & 53.33 & 39.41 & 39.41 & 3.33 & 6.67 & 7.56 & 7.56 \\ \cmidrule{1-1} \cmidrule{3-16} 
\multicolumn{1}{c|}{soft} & \multicolumn{1}{c|}{} & \multicolumn{1}{c|}{} & log & 13.33 & 16.67 & 10.49 & 10.49 & 30.00 & 40.00 & 15.58 & 15.58 & 3.33 & 3.33 & 3.22 & 3.22 \\
\multicolumn{1}{c|}{soft} & \multicolumn{1}{c|}{} & \multicolumn{1}{c|}{} & NLL & 10.00 & 16.67 & 11.80 & 11.80 & 3.33 & 10.00 & 6.84 & 6.84 & 3.33 & 3.33 & 4.01 & 4.01 \\
\multicolumn{1}{c|}{manual} & \multicolumn{1}{c|}{} & \multicolumn{1}{c|}{\multirow{-3}{*}{max}} & - & 40.00 & 46.67 & 20.30 & 20.30 & 43.33 & 50.00 & 19.99 & 19.99 & 0.00 & 0.00 & 0.81 & 0.81 \\ \cmidrule{1-1} \cmidrule{3-16} 
\multicolumn{1}{c|}{soft} & \multicolumn{1}{c|}{} & \multicolumn{1}{c|}{} & log & 20.00 & 60.00 & 39.13 & 39.13 & 10.00 & 56.67 & 29.65 & 29.65 & 43.33 & 53.33 & 48.18 & 48.18 \\
\multicolumn{1}{c|}{soft} & \multicolumn{1}{c|}{} & \multicolumn{1}{c|}{} & NLL & 20.00 & 56.67 & 39.01 & 39.01 & 6.67 & 50.00 & 25.47 & 25.47 & 20.00 & 56.67 & 36.26 & 36.26 \\
\multicolumn{1}{c|}{manual} & \multicolumn{1}{c|}{\multirow{-9}{*}{s}} & \multicolumn{1}{c|}{\multirow{-3}{*}{mean}} & - & 43.33 & 53.33 & 47.63 & 47.63 & 43.33 & 53.33 & 49.34 & 49.34 & 6.67 & 20.00 & 15.31 & 15.31 \\ \midrule
 &  &  &  & \multicolumn{4}{c|}{BERT-LARGE-CASED} & \multicolumn{4}{c|}{BERT-LARGE-UNCASED} & \multicolumn{4}{c}{RoBERTa-LARGE} \\ \midrule
\multicolumn{1}{c|}{soft} & \multicolumn{1}{c|}{} & \multicolumn{1}{c|}{} & log & 40.00 & 60.00 & 48.67 & 48.67 & 0.00 & 6.67 & 3.77 & 3.77 & 6.67 & 13.33 & 10.23 & 10.23 \\
\multicolumn{1}{c|}{soft} & \multicolumn{1}{c|}{} & \multicolumn{1}{c|}{} & NLL & 16.67 & 30.00 & 24.71 & 24.71 & 26.67 & 40.00 & 33.48 & 33.48 & 13.33 & 16.67 & 15.05 & 15.05 \\
\multicolumn{1}{c|}{manual} & \multicolumn{1}{c|}{} & \multicolumn{1}{c|}{\multirow{-3}{*}{first}} & - & 33.33 & 50.00 & 42.28 & 42.28 & 30.00 & 46.67 & 39.19 & 39.19 & 0.00 & 0.00 & 3.75 & 3.75 \\ \cmidrule{1-1} \cmidrule{3-16} 
\multicolumn{1}{c|}{soft} & \multicolumn{1}{c|}{} & \multicolumn{1}{c|}{} & log & 23.33 & 33.33 & 29.60 & 29.60 & 13.33 & 26.67 & 19.89 & 19.89 & 0.00 & 13.33 & 5.41 & 5.41 \\
\multicolumn{1}{c|}{soft} & \multicolumn{1}{c|}{} & \multicolumn{1}{c|}{} & NLL & 20.00 & 43.33 & 29.44 & 29.44 & 6.67 & 13.33 & 10.59 & 10.59 & 0.00 & 0.00 & 0.60 & 0.60 \\
\multicolumn{1}{c|}{manual} & \multicolumn{1}{c|}{} & \multicolumn{1}{c|}{\multirow{-3}{*}{max}} & - & 33.33 & 43.33 & 38.52 & 38.52 & 23.33 & 36.67 & 30.94 & 30.94 & 0.00 & 0.00 & 0.36 & 0.36 \\ \cmidrule{1-1} \cmidrule{3-16} 
\multicolumn{1}{c|}{soft} & \multicolumn{1}{c|}{} & \multicolumn{1}{c|}{} & log & 26.67 & 56.67 & 39.09 & 39.09 & 6.67 & 26.67 & 15.42 & 15.42 & 13.33 & 30.00 & 23.08 & 23.08 \\
\multicolumn{1}{c|}{soft} & \multicolumn{1}{c|}{} & \multicolumn{1}{c|}{} & NLL & 36.67 & 63.33 & 48.11 & 48.11 & 6.67 & 13.33 & 10.15 & 10.15 & 10.00 & 50.00 & 25.53 & 25.53 \\
\multicolumn{1}{c|}{manual} & \multicolumn{1}{c|}{\multirow{-9}{*}{m}} & \multicolumn{1}{c|}{\multirow{-3}{*}{mean}} & - & 46.67 & 50.00 & 50.33 & 50.33 & 30.00 & 53.33 & 41.43 & 41.43 & 0.00 & 0.00 & 1.73 & 1.73 \\ \midrule
\multicolumn{1}{c|}{soft} & \multicolumn{1}{c|}{} & \multicolumn{1}{c|}{} & log & 30.00 & 56.67 & 34.30 & 34.30 & 6.67 & 16.67 & 13.42 & 13.42 & 10.00 & 13.33 & 14.15 & 14.15 \\
\multicolumn{1}{c|}{soft} & \multicolumn{1}{c|}{} & \multicolumn{1}{c|}{} & NLL & 16.67 & 46.67 & 30.00 & 30.00 & 26.67 & 50.00 & 32.20 & 32.20 & 6.67 & 6.67 & 8.50 & 8.50 \\
\multicolumn{1}{c|}{manual} & \multicolumn{1}{c|}{} & \multicolumn{1}{c|}{\multirow{-3}{*}{first}} & - & 40.00 & 50.00 & 30.83 & 30.83 & 33.33 & 50.00 & 32.08 & 32.08 & 13.33 & 46.67 & 27.36 & 27.36 \\ \cmidrule{1-1} \cmidrule{3-16} 
\multicolumn{1}{c|}{soft} & \multicolumn{1}{c|}{} & \multicolumn{1}{c|}{} & log & 10.00 & 23.33 & 11.55 & 11.55 & 6.67 & 6.67 & 8.69 & 8.69 & 0.00 & 0.00 & 0.34 & 0.34 \\
\multicolumn{1}{c|}{soft} & \multicolumn{1}{c|}{} & \multicolumn{1}{c|}{} & NLL & 23.33 & 36.67 & 16.67 & 16.67 & 6.67 & 10.00 & 6.69 & 6.69 & 16.67 & 20.00 & 17.79 & 17.79 \\
\multicolumn{1}{c|}{manual} & \multicolumn{1}{c|}{} & \multicolumn{1}{c|}{\multirow{-3}{*}{max}} & - & 40.00 & 50.00 & 18.87 & 18.87 & 30.00 & 43.33 & 16.04 & 16.04 & 0.00 & 3.33 & 1.31 & 1.31 \\ \cmidrule{1-1} \cmidrule{3-16} 
\multicolumn{1}{c|}{soft} & \multicolumn{1}{c|}{} & \multicolumn{1}{c|}{} & log & 30.00 & 53.33 & 40.98 & 40.98 & 20.00 & 46.67 & 32.22 & 32.22 & 0.00 & 10.00 & 5.54 & 5.54 \\
\multicolumn{1}{c|}{soft} & \multicolumn{1}{c|}{} & \multicolumn{1}{c|}{} & NLL & 26.67 & 53.33 & 38.80 & 38.80 & 6.67 & 10.00 & 9.26 & 9.26 & 10.00 & 23.33 & 18.06 & 18.06 \\
\multicolumn{1}{c|}{manual} & \multicolumn{1}{c|}{\multirow{-9}{*}{s}} & \multicolumn{1}{c|}{\multirow{-3}{*}{mean}} & - & 46.67 & 50.00 & 50.18 & 50.18 & 33.33 & 53.33 & 43.17 & 43.17 & 3.33 & 16.67 & 11.58 & 11.58 \\ \bottomrule[1.5pt]
\end{tabular}%
}
\caption{DM results.}
\label{DM}
\end{table*}

\begin{table*}[htbp]
\scalebox{0.7}{%
\begin{tabular}{cccc|cccc|cccc|cccc}
\toprule[1.5pt]
 &  &  &  & \multicolumn{4}{c|}{BERT-BASE-CASED} & \multicolumn{4}{c|}{BERT-BASE-UNCASED} & \multicolumn{4}{c}{RoBERTa-BASE} \\ \midrule
\multicolumn{1}{c|}{Template} & \multicolumn{1}{c|}{Masks} & \multicolumn{1}{c|}{Pooling} & Loss & R@1 & R@5 & MRRa & MRR & R@1 & R@5 & MRRa & MRR & R@1 & R@5 & MRRa & MRR \\ \midrule
\multicolumn{1}{c|}{soft} & \multicolumn{1}{c|}{} & \multicolumn{1}{c|}{} & log & 42.86 & 53.57 & 51.49 & 51.49 & 46.43 & 67.86 & 55.89 & 55.89 & 39.29 & 53.57 & 44.34 & 44.34 \\
\multicolumn{1}{c|}{soft} & \multicolumn{1}{c|}{} & \multicolumn{1}{c|}{} & NLL & 53.57 & 60.71 & 58.18 & 58.18 & 46.43 & 64.29 & 55.87 & 55.87 & 32.14 & 50 & 40.67 & 40.67 \\
\multicolumn{1}{c|}{manual} & \multicolumn{1}{c|}{} & \multicolumn{1}{c|}{\multirow{-3}{*}{first}} & - & 39.29 & 60.71 & 48.89 & 48.89 & 28.57 & 67.86 & 44.74 & 44.74 & 10.71 & 39.29 & 22.7 & 22.7 \\ \cmidrule{1-1} \cmidrule{3-16} 
\multicolumn{1}{c|}{soft} & \multicolumn{1}{c|}{} & \multicolumn{1}{c|}{} & log & 35.71 & 57.14 & 45.6 & 45.6 & 39.29 & 71.43 & 53.1 & 53.1 & 17.86 & 46.43 & 30.01 & 30.01 \\
\multicolumn{1}{c|}{soft} & \multicolumn{1}{c|}{} & \multicolumn{1}{c|}{} & NLL & 17.86 & 50 & 34.58 & 34.58 & 42.86 & 50 & 47.38 & 47.38 & 39.29 & 46.43 & 43.27 & 43.27 \\
\multicolumn{1}{c|}{manual} & \multicolumn{1}{c|}{} & \multicolumn{1}{c|}{\multirow{-3}{*}{max}} & - & 46.43 & 57.14 & 48.64 & 48.64 & 32.14 & 57.14 & 39.43 & 39.43 & 0 & 0 & 1.23 & 1.23 \\ \cmidrule{1-1} \cmidrule{3-16} 
\multicolumn{1}{c|}{soft} & \multicolumn{1}{c|}{} & \multicolumn{1}{c|}{} & log & 42.86 & 60.71 & 50.87 & 50.87 & 46.43 & 75 & 58.74 & 58.74 & 46.43 & 50 & 50.32 & 50.32 \\
\multicolumn{1}{c|}{soft} & \multicolumn{1}{c|}{} & \multicolumn{1}{c|}{} & NLL & 50 & 67.86 & 57.81 & 57.81 & 46.43 & 67.86 & 57.86 & 57.86 & 39.29 & 53.57 & 47.69 & 47.69 \\
\multicolumn{1}{c|}{manual} & \multicolumn{1}{c|}{\multirow{-9}{*}{m}} & \multicolumn{1}{c|}{\multirow{-3}{*}{mean}} & - & 46.43 & 67.86 & 59.08 & 59.08 & 42.86 & 75 & 55.97 & 55.97 & 14.29 & 32.14 & 23.52 & 23.52 \\ \midrule
\multicolumn{1}{c|}{soft} & \multicolumn{1}{c|}{} & \multicolumn{1}{c|}{} & log & 46.43 & 60.71 & 53.94 & 53.94 & 42.86 & 60.71 & 41.7 & 41.7 & 35.71 & 46.43 & 38.4 & 38.4 \\
\multicolumn{1}{c|}{soft} & \multicolumn{1}{c|}{} & \multicolumn{1}{c|}{} & NLL & 50 & 64.29 & 52.59 & 52.59 & 42.86 & 67.86 & 39.37 & 39.37 & 28.57 & 53.57 & 38.78 & 38.78 \\
\multicolumn{1}{c|}{manual} & \multicolumn{1}{c|}{} & \multicolumn{1}{c|}{\multirow{-3}{*}{first}} & - & 42.86 & 60.71 & 42.73 & 42.73 & 39.29 & 64.29 & 35.56 & 35.56 & 14.29 & 35.71 & 25.13 & 25.13 \\ \cmidrule{1-1} \cmidrule{3-16} 
\multicolumn{1}{c|}{soft} & \multicolumn{1}{c|}{} & \multicolumn{1}{c|}{} & log & 39.29 & 53.57 & 26.05 & 26.05 & 35.71 & 57.14 & 23.6 & 23.6 & 32.14 & 35.71 & 34.04 & 34.04 \\
\multicolumn{1}{c|}{soft} & \multicolumn{1}{c|}{} & \multicolumn{1}{c|}{} & NLL & 39.29 & 50 & 27.73 & 27.73 & 42.86 & 46.43 & 26.98 & 26.98 & 32.14 & 42.86 & 36.1 & 36.1 \\
\multicolumn{1}{c|}{manual} & \multicolumn{1}{c|}{} & \multicolumn{1}{c|}{\multirow{-3}{*}{max}} & - & 42.86 & 60.71 & 30.76 & 30.76 & 42.86 & 53.57 & 24.73 & 24.73 & 3.57 & 3.57 & 4.78 & 4.78 \\ \cmidrule{1-1} \cmidrule{3-16} 
\multicolumn{1}{c|}{soft} & \multicolumn{1}{c|}{} & \multicolumn{1}{c|}{} & log & 57.14 & 64.29 & 62.72 & 62.72 & 57.14 & 71.43 & 63.93 & 63.93 & 39.29 & 53.57 & 46.41 & 46.41 \\
\multicolumn{1}{c|}{soft} & \multicolumn{1}{c|}{} & \multicolumn{1}{c|}{} & NLL & 46.43 & 67.86 & 56.19 & 56.19 & 53.57 & 71.43 & 62.39 & 62.39 & 32.14 & 53.57 & 42.03 & 42.03 \\
\multicolumn{1}{c|}{manual} & \multicolumn{1}{c|}{\multirow{-9}{*}{s}} & \multicolumn{1}{c|}{\multirow{-3}{*}{mean}} & - & 42.86 & 67.86 & 56.61 & 56.61 & 39.29 & 75 & 53.48 & 53.48 & 32.14 & 57.14 & 43.97 & 43.97 \\ \midrule
 &  &  &  & \multicolumn{4}{c|}{BERT-LARGE-CASED} & \multicolumn{4}{c|}{BERT-LARGE-UNCASED} & \multicolumn{4}{c}{RoBERTa-LARGE} \\ \midrule
\multicolumn{1}{c|}{soft} & \multicolumn{1}{c|}{} & \multicolumn{1}{c|}{} & log & 57.14 & 75 & 66.24 & 66.24 & 0 & 0 & 1.53 & 1.53 & 35.71 & 60.71 & 46.25 & 46.25 \\
\multicolumn{1}{c|}{soft} & \multicolumn{1}{c|}{} & \multicolumn{1}{c|}{} & NLL & 53.57 & 67.86 & 61.64 & 61.64 & 35.71 & 67.86 & 49.25 & 49.25 & 28.57 & 67.86 & 44.01 & 44.01 \\
\multicolumn{1}{c|}{manual} & \multicolumn{1}{c|}{} & \multicolumn{1}{c|}{\multirow{-3}{*}{first}} & - & 46.43 & 67.86 & 56.92 & 56.92 & 39.29 & 71.43 & 51.73 & 51.73 & 0 & 14.29 & 8.29 & 8.29 \\ \cmidrule{1-1} \cmidrule{3-16} 
\multicolumn{1}{c|}{soft} & \multicolumn{1}{c|}{} & \multicolumn{1}{c|}{} & log & 50 & 75 & 60.48 & 60.48 & 35.71 & 57.14 & 44.92 & 44.92 & 25 & 32.14 & 28.94 & 28.94 \\
\multicolumn{1}{c|}{soft} & \multicolumn{1}{c|}{} & \multicolumn{1}{c|}{} & NLL & 17.86 & 60.71 & 37.55 & 37.55 & 28.57 & 67.86 & 41.38 & 41.38 & 0 & 3.57 & 1.51 & 1.51 \\
\multicolumn{1}{c|}{manual} & \multicolumn{1}{c|}{} & \multicolumn{1}{c|}{\multirow{-3}{*}{max}} & - & 46.43 & 57.14 & 52.08 & 52.08 & 39.29 & 60.71 & 46.45 & 46.45 & 0 & 0 & 1.01 & 1.01 \\ \cmidrule{1-1} \cmidrule{3-16} 
\multicolumn{1}{c|}{soft} & \multicolumn{1}{c|}{} & \multicolumn{1}{c|}{} & log & 53.57 & 67.86 & 60.88 & 60.88 & 28.57 & 46.43 & 39.13 & 39.13 & 35.71 & 57.14 & 45.3 & 45.3 \\
\multicolumn{1}{c|}{soft} & \multicolumn{1}{c|}{} & \multicolumn{1}{c|}{} & NLL & 50 & 71.43 & 60.42 & 60.42 & 46.43 & 75 & 59.46 & 59.46 & 32.14 & 67.86 & 46.06 & 46.06 \\
\multicolumn{1}{c|}{manual} & \multicolumn{1}{c|}{\multirow{-9}{*}{m}} & \multicolumn{1}{c|}{\multirow{-3}{*}{mean}} & - & 57.14 & 78.57 & 66.82 & 66.82 & 46.43 & 78.57 & 61.06 & 61.06 & 7.14 & 17.86 & 14.74 & 14.74 \\ \midrule
\multicolumn{1}{c|}{soft} & \multicolumn{1}{c|}{} & \multicolumn{1}{c|}{} & log & 46.43 & 60.71 & 45.3 & 45.3 & 32.14 & 60.71 & 38.88 & 38.88 & 42.86 & 53.57 & 44.55 & 44.55 \\
\multicolumn{1}{c|}{soft} & \multicolumn{1}{c|}{} & \multicolumn{1}{c|}{} & NLL & 39.29 & 71.43 & 51.73 & 51.73 & 53.57 & 67.86 & 44.62 & 44.62 & 25 & 28.57 & 26.3 & 26.3 \\
\multicolumn{1}{c|}{manual} & \multicolumn{1}{c|}{} & \multicolumn{1}{c|}{\multirow{-3}{*}{first}} & - & 50 & 67.86 & 53.54 & 53.54 & 42.86 & 71.43 & 41.39 & 41.39 & 10.71 & 53.57 & 29.46 & 29.46 \\ \cmidrule{1-1} \cmidrule{3-16} 
\multicolumn{1}{c|}{soft} & \multicolumn{1}{c|}{} & \multicolumn{1}{c|}{} & log & 42.86 & 64.29 & 31.5 & 31.5 & 25 & 42.86 & 18.64 & 18.64 & 17.86 & 42.86 & 27.04 & 27.04 \\
\multicolumn{1}{c|}{soft} & \multicolumn{1}{c|}{} & \multicolumn{1}{c|}{} & NLL & 42.86 & 57.14 & 28.5 & 28.5 & 0 & 7.14 & 2.26 & 2.26 & 25 & 50 & 32.22 & 32.22 \\
\multicolumn{1}{c|}{manual} & \multicolumn{1}{c|}{} & \multicolumn{1}{c|}{\multirow{-3}{*}{max}} & - & 42.86 & 67.86 & 32.89 & 32.89 & 46.43 & 53.57 & 31.91 & 31.91 & 3.57 & 7.14 & 2.35 & 2.35 \\ \cmidrule{1-1} \cmidrule{3-16} 
\multicolumn{1}{c|}{soft} & \multicolumn{1}{c|}{} & \multicolumn{1}{c|}{} & log & 0 & 0 & 1.71 & 1.71 & 46.43 & 64.29 & 55.81 & 55.81 & 32.14 & 60.71 & 43.23 & 43.23 \\
\multicolumn{1}{c|}{soft} & \multicolumn{1}{c|}{} & \multicolumn{1}{c|}{} & NLL & 42.86 & 67.86 & 56.91 & 56.91 & 35.71 & 53.57 & 45.75 & 45.75 & 35.71 & 71.43 & 48.53 & 48.53 \\
\multicolumn{1}{c|}{manual} & \multicolumn{1}{c|}{\multirow{-9}{*}{s}} & \multicolumn{1}{c|}{\multirow{-3}{*}{mean}} & - & 57.14 & 75 & 66.62 & 66.62 & 42.86 & 78.57 & 58.3 & 58.3 & 17.86 & 50 & 33.22 & 33.22 \\ \bottomrule[1.5pt]
\end{tabular}%
}
\caption{RG results.}
\label{RG}
\end{table*}

\subsection{Reasoning Results}
\begin{figure*}[ht]
    \centering
    \includegraphics[width=\textwidth]{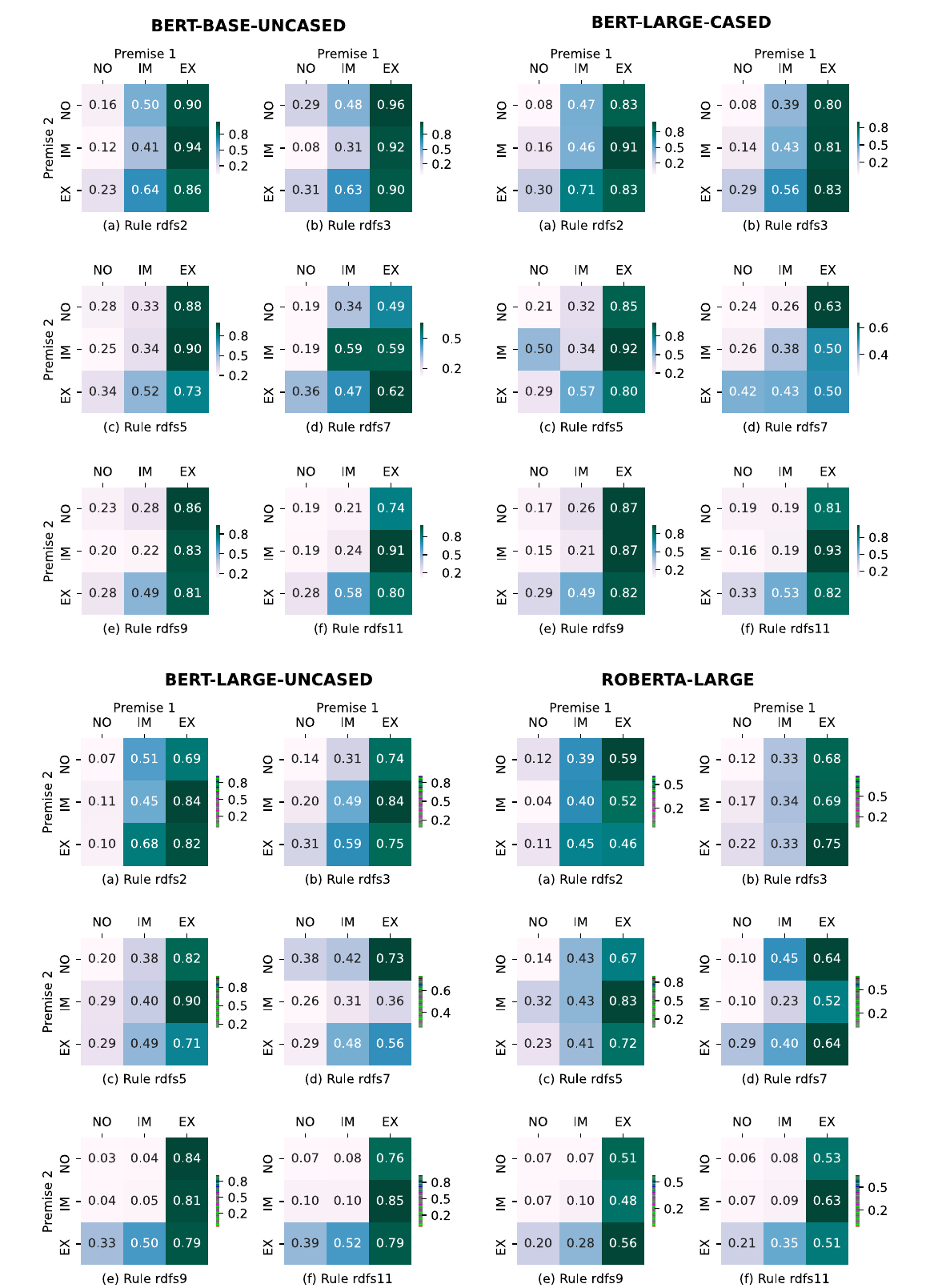}
    \caption{The MRR by BERT-base-uncased, BERT-large-(un)cased and RoBERTa-large using different combinations of premises. EX stands for explicitly given, IM stands for implicitly given and NO stands for not given. The other metrics show similar trends.}
    \label{reasoning_all}
\end{figure*}

We report the MRR Metric of BERT-base-uncased, BERT-large-cased, BERT-large-uncased and RoBERTa-large in Figure~\ref{reasoning_all}. It is generally consistent with the two models reported in the main body of the paper and the macro-averaged performance across different PLMs, so consistent conclusions can be drawn.

\end{document}